\RequirePackage{xcolor}
\documentclass[journal]{IEEEtran}
\usepackage{cite}
\usepackage{amsmath,amssymb,amsfonts}
\usepackage{algorithmic}
\usepackage{graphicx}
\usepackage{textcomp}
\usepackage{booktabs}
\usepackage{multirow}
\usepackage{colortbl}
\usepackage{pifont}
\usepackage{makecell}
\usepackage[ruled,linesnumbered]{algorithm2e}
\usepackage{hyperref}
\hypersetup{hidelinks,
	colorlinks=true,
	allcolors=black,
	pdfstartview=Fit,
	breaklinks=true}
\def\BibTeX{{\rm B\kern-.05em{\sc i\kern-.025em b}\kern-.08em
    T\kern-.1667em\lower.7ex\hbox{E}\kern-.125emX}}
\begin{document}
%\title{Mixup Augmentation with Pseudo-Bags for Weakly-Supervised Whole Slide Image Classification}
\title{Pseudo-Bag Mixup Augmentation for Multiple Instance Learning-Based Whole Slide Image Classification}
\author{Pei Liu, Luping Ji \IEEEmembership{Member, IEEE}, Xinyu Zhang, and Feng Ye
%\thanks{Manuscript received xx xx, xxxx. This work is supported by the Fundamental Research Funds for the Central Universities, ZYGX2022YGRH015, partly by the National Natural Science Foundation of China (NSFC) under Grant No. 61972072.}
\thanks{Pei Liu, Luping Ji, and Xinyu Zhang are with the School of Computer Science and Engineering, University of Electronic Science and Technology of China, Chengdu 611731, China.}
\thanks{Feng Ye is with the Institute of Clinical Pathology, West China Hospital, Sichuan University, Guo Xue Xiang, Chengdu 610041, China.}
\thanks{* Corresponding author: Luping Ji (jiluping@uestc.edu.cn).}
}

\maketitle

\begin{abstract}
Given the special situation of modeling gigapixel images, multiple instance learning (MIL) has become one of the most important frameworks for Whole Slide Image (WSI) classification. 
In current practice, most MIL networks often face two unavoidable problems in training: i) insufficient WSI data and ii) the sample memorization inclination inherent in neural networks. 
These problems may hinder MIL models from adequate and efficient training, suppressing the continuous performance promotion of classification models on WSIs. 
Inspired by the basic idea of Mixup, this paper proposes a new Pseudo-bag Mixup (PseMix) data augmentation scheme to improve the training of MIL models. 
This scheme generalizes the Mixup strategy for general images to special WSIs via pseudo-bags so as to be applied in MIL-based WSI classification. 
Cooperated by pseudo-bags, our PseMix fulfills the critical size alignment and semantic alignment in Mixup strategy. 
Moreover, it is designed as an efficient and decoupled method, neither involving time-consuming operations nor relying on MIL model predictions.
Comparative experiments and ablation studies are specially designed to evaluate the effectiveness and advantages of our PseMix. Experimental results show that PseMix could often assist state-of-the-art MIL networks to refresh their classification performance on WSIs.
Besides, it could also boost the generalization performance of MIL models in special test scenarios, and promote their robustness to patch occlusion and label noise. 
Our source code is available at \href{https://github.com/liupei101/PseMix}{\textcolor[RGB]{0,67,147}{https://github.com/liupei101/PseMix}}.
%Although it has witnessed remarkable success, those frequently-used MIL networks, similar to traditional neural networks, are prone to memorize training data rather than generalize from it. Such a behavior is even more obvious in the WSI data currently at a small scale. However, previous studies rarely investigate it in a formal way. As a result, the behavior of data memorization still greatly limits the potential of MIL models and impedes their development for clinical applications. 
%However, an empirical study reveals that those frequently-used MIL networks are prone to a gradual decline in generalization during training, mainly caused by insufficient training data and the notorious data memorization behavior of neural networks (NNs).This problem could significantly harm the performance of MIL models on new WSI data, thus calling into question their clinical usability. 
%The proposed PseMix could serve as a promising data augmentation scheme to help develop the MIL models with better generalization and robustness for pathology diagnosis. 
\end{abstract}


\section{Introduction}
\label{sec:introduction}

Histological Whole-Slide Image (WSI) serves as the gold standard of pathology diagnosis and plays a vital role in cancer assessment and treatment \cite{Bera2019,Niazi2019,Zarella2019}. 
Owing to the advances in deep learning, many clinical tasks with WSIs, such as lymph node metastasis detection, cancer subtyping, and cancer prognosis, have been made automatic and precise by specialized models \cite{Mercan8055584,Campanella2019,shao2021weakly,Lu2021,li2021dual,shao2021transmil,Wang9765508,zhang2022dtfd,liu2023advmil,Zhu2023,Zheng2023,Shi10119190}. 
These models show exciting progress towards overcoming the long-standing drawbacks of manual inspection \cite{Koohbanani9343323,Wang9869889}. 
Across them, a weakly-supervised learning framework called multiple instance learning (MIL) is widely used. It treats a gigapixel WSI (\textit{e.g.}, 40,000 $\times$ 40,000 pixels) as a bag of multiple instances, and learns a global feature representation via instance embedding and aggregation \cite{ilse2018attn}. This framework enables neural network (NN) models to handle gigapixel images and no longer rely on fine-grained instance-level labels. For those reasons, MIL has become a prevalent paradigm for WSI classification, achieving considerable successes in computational pathology (CPATH) \cite{Chen2022}.

\begin{figure}[tp]
\centering
\includegraphics[width=\columnwidth]{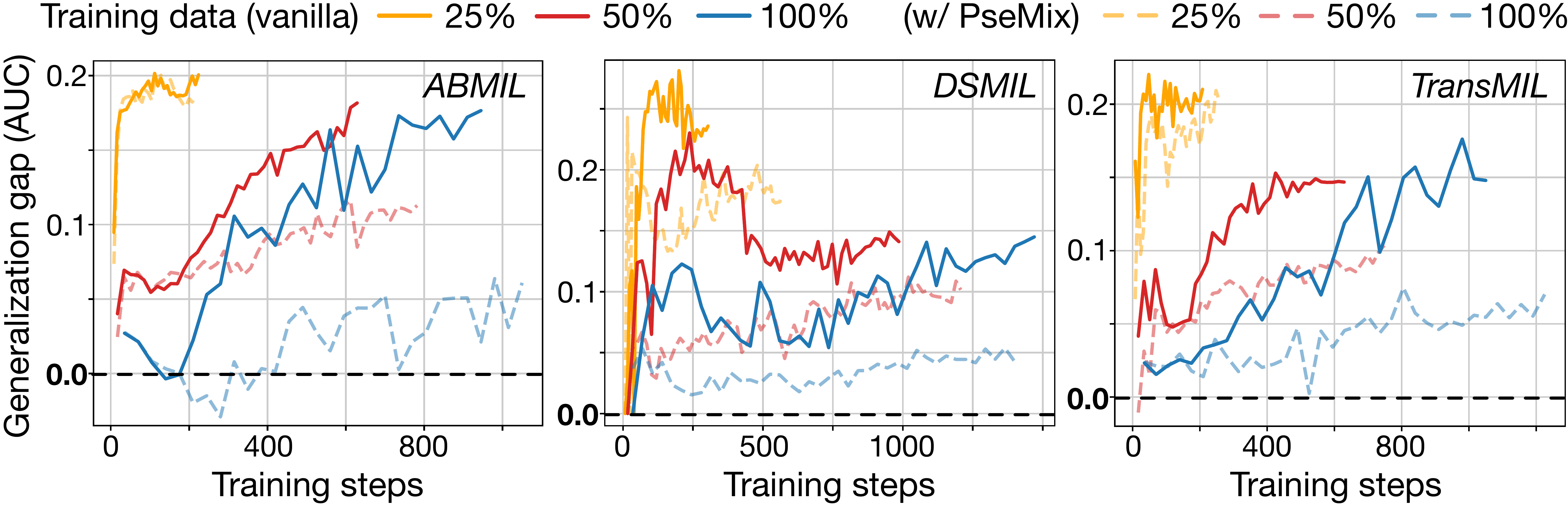}
\caption{Generalization gap of MIL models in training. It is measured by the model performance gap between training and test sets, to assess the generalization ability of models, following \cite{jiang2018predicting,zhang2021how}. Three state-of-the-art MIL models are trained on TCGA-BRCA WSIs. Vanilla models often show rapidly-growing gaps. The models trained with our PseMix could ease these without introducing extra complicated techniques.}
\label{fig:intro-gen-gap}
\end{figure}

Nonetheless, our empirical study, as presented in Figure \ref{fig:intro-gen-gap}, reveals that some state-of-the-art MIL models \cite{ilse2018attn,li2021dual,shao2021transmil} often perform poorly in generalization, even worse in probable small data scenarios. %Moreover, that performance would be worse when training them with less WSI samples.
The literatures on learning theory and NNs \cite{Vapnik2000,szegedy2014intriguing,zhang2017understanding} point out that such kind of problem is the consequence of \textit{data memorization}, fundamentally caused by the nature of NNs and existing even in the presence of strong regularization. 
Due to internal data memorization characteristics, MIL networks could merely memorize, rather than efficiently learn or generalize from, given training samples, thus limiting the model performance on unseen test data and increasing generalization gaps. Besides, as an external cause, insufficient samples may hinder models from inadequate training, further degrading generalization performance. 
In contrast to a multitude of studies on classical image classification for tackling the data memorization problem in models \cite{jiang2018predicting,Li9302891}, the current works on WSI classification rarely investigate this problem in MIL formally. Thereby, most modern MIL models still carry the risk of inadequate and inefficient training, suppressing their continuous performance promotion. 

Mixup \cite{zhang2018mixup}, as a special data augmentation method, shows great promise in mitigating the inherent data memorization behavior in NNs. Unlike most augmentation methods that only manipulate single input, Mixup operates on two inputs and generates the interpolation samples between them ($\widetilde{x} = \lambda x_i + (1-\lambda) x_j,\ \widetilde{y} = \lambda y_i + (1-\lambda) y_j$). These samples are also interpreted as neighbor or vicinity samples, leveraged as augmented data for training. In this particular way, NNs are never limited to memorizing finite training samples from the original distribution; instead, they are encouraged to learn from the vicinity distribution of training samples and obtain a decision boundary with better generalization \cite{zhang2018mixup}. 
Such a simple yet efficient method has been developed and witnessed remarkable success in its application to heterogeneous data, \textit{e.g.}, natural image \cite{yun9008296,pmlr-v119-kim20b,chen9879590}, graph \cite{pmlr-v162-han22c}, 3D point cloud, \cite{wang2023pointpatchmix}, visual-language \cite{Gong9802503}, etc. 
Further adoption of Mixup for histological WSIs is strongly anticipated. However, the characteristic of WSIs poses several challenges to it. Unlike other data sources, i) histological WSIs are usually cast as bags for weakly-supervised MIL; ii) WSI bags are often irregular and not well-aligned in their representation space for interpolation (\textit{i.e.}, $\widetilde{x} = \lambda x_i + (1-\lambda) x_j$); and iii) a single bag may contain tens of thousands of instances, each with a high-dimensional feature vector \cite{Lu2021}. These characteristics suggest that a practical and suitable Mixup method for WSIs should be compatible with MIL, capable of mixing alignment, and efficient. 

%Through comparative analyses and induction, we find that most successful Mixup variants, such as TransMix  \cite{chen9879590} and $\mathcal{G}$-Mixup \cite{pmlr-v162-han22c}, though varied in designs, share two commonalities: i) \textit{size alignment}, \textit{i.e.}, $x_i$ and $x_j$ are forced to be aligned in Euclidean space before mixing, and ii) \textit{semantic alignment}, \textit{i.e.}, the mixed input $\widetilde{x}$ and the mixed label $\widetilde{y}$ are aligned in semantics via the same proportion of mixture, as shown in Figure \ref{fig:intro-mixup} (refer to Table \ref{tb0} for more details). 

\begin{figure}[tp]
\centering
\includegraphics[width=0.78\columnwidth]{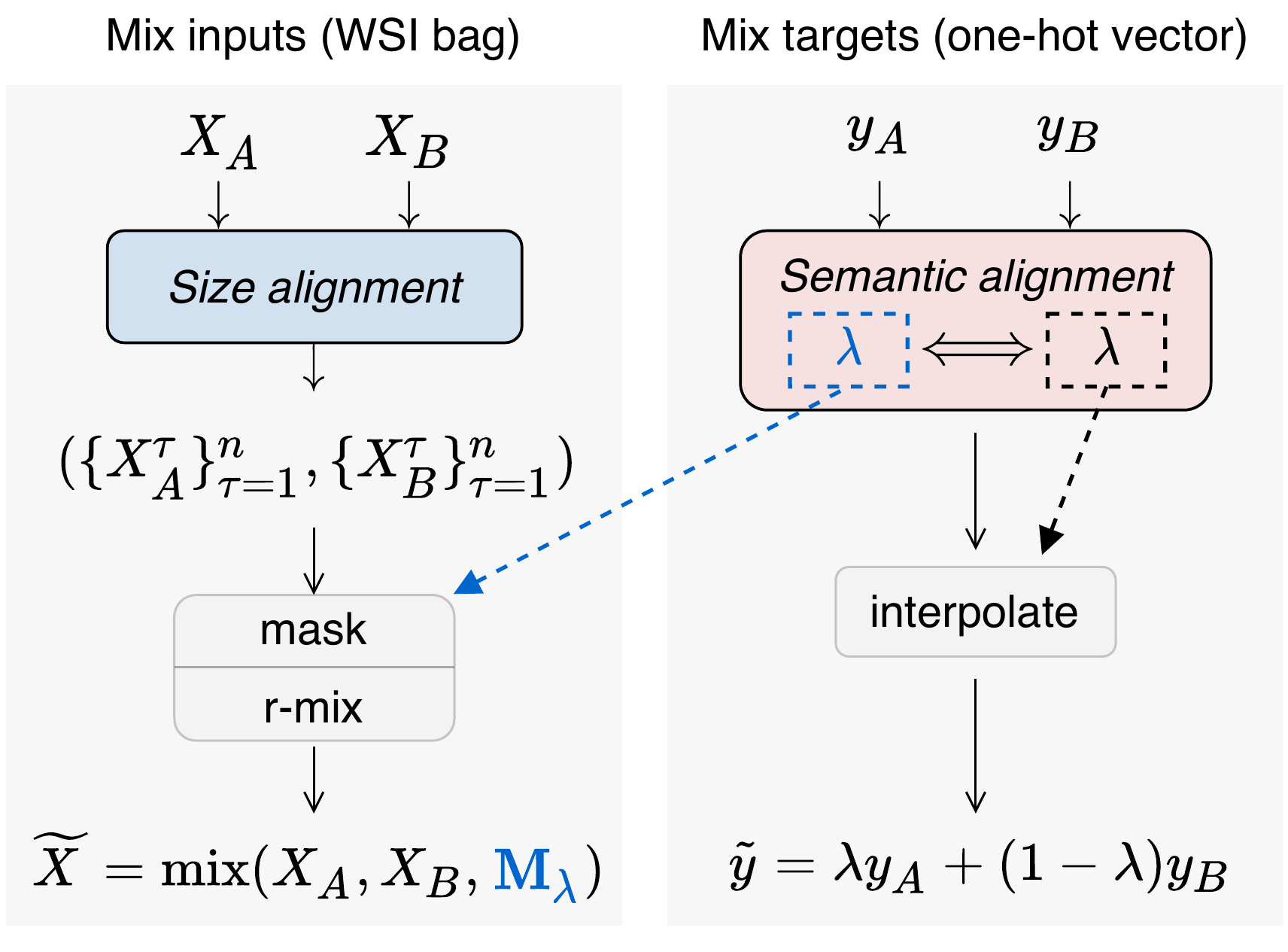}
\caption{A conceptual framework of PseMix for MIL-based WSI classification. Size alignment means that two input representations (\textit{e.g.}, feature vectors and image matrices) are aligned in size at every dimension involving mixing. 
$X_{A}^{\tau}$ and $X_{B}^{\tau}$ denote pseudo-bags. $\mathbf{M}_{\lambda}$ is a binary mask. R-mix means random mixing. PseMix generalizes Mixup and fulfills its critical size alignment and semantic alignment via pseudo-bags.}
\label{fig:intro-mixup}
\end{figure}

For the purposes above, this paper proposes a new Pseudo-bag Mixup (PseMix) data augmentation scheme, as illustrated in Figure \ref{fig:intro-mixup}. We address the challenges aforementioned from the following three aspects. (1) \textit{MIL compatibility}: our scheme takes WSI bags as input and outputs the mixed bags that still could be processed by MIL. (2) \textit{Mixing alignment}: we divide each bag into $n$ pseudo-bags for size alignment in bag mixing, and leverage the proportion of pseudo-bag mixing as $\lambda$ to mix targets for semantic alignment. %share the $\lambda$ between bag mixing and target mixing for semantic alignment. 
(3) \textit{Efficiency}: our pseudo-bag division is designed as an algorithm with linear time and space complexity, and PseMix is decoupled from the stage of MIL so as to be plugin-and-play. In addition, our PseMix further introduces a random mixing mechanism (\textit{i.e.}, r-mix in Figure \ref{fig:intro-mixup}) for more data diversity and efficient learning on vicinity samples. 
Comparative experiments confirm that our PseMix could serve as an effective Mixup data augmentation method. Apart from the improvements across three different classification tasks, the models trained with our PseMix could often surpass vanilla models by a large margin in various generalization and robustness tests.

The main contributions of this paper are as follows. (1) This paper proposes a new Pseudo-bag Mixup (PseMix) data augmentation scheme for MIL-based WSI classification, to help MIL models improve performance and obtain better generalization and robustness. It is an efficient and plugin-and-play one, neither involving time-consuming operations nor relying on the prediction of MIL models. (2) This paper generalizes the Mixup strategy for common images to special WSIs by pseudo-bag-based size and semantic alignment. %the critical size and semantic alignment on pseudo-bags. 
Furthermore, it introduces a random mixing mechanism into standard Mixup for more data diversity and efficient learning on augmented mixing samples. (3) Comparative experiments, ablation study, and various scenarios, \textit{e.g.}, regular WSI classification, special generalization tests, patch occlusion robustness, label noise robustness, etc, are specially designed to demonstrate the broad advantages of PseMix.

\section{Related work}

\begin{table*}[tp]
\small
\centering
\caption{Related data augmentation methods. The mix manner, r-mix, means random mixing.}
\label{tb0}
\begin{tabular}{l|l|cll|cl}
\toprule
\multirow{2}{*}{Method} & \multirow{2}{*}{Data type} & \multicolumn{3}{c|}{Input} & \multicolumn{2}{c}{Target}   \\ %\cmidrule(lr){3-5} \cmidrule(lr){6-7}
 & & Mix & Size alignment & Mix manner & Mix & Semantic alignment \\
\midrule

Mixup (2018) \cite{zhang2018mixup}    & data-agnostic &  \ding{51} &  aligned by default & interpolate  & \ding{51} & input interpolation scale\\
TransMix (2022) \cite{chen9879590} & natural image         & \ding{51}   & scaling or cropping & mask \& mix & \ding{51} &  target attention weights   \\
$\mathcal{G}$-Mixup (2022) \cite{pmlr-v162-han22c} & graph &  \ding{51}      &  graphon estimation & interpolate & \ding{51} & graphon interpolation scale   \\
PointPatchMix (2023) \cite{wang2023pointpatchmix} & point cloud & \ding{51}  & point patch & mask \& mix & \ding{51} &  patch attention scores  \\ \cmidrule(lr){1-7}
Pseudo-bag (2021) \cite{shao2021weakly} & WSI & \ding{55} & ---  & --- & \ding{55} & --- \\
ReMix (2022) \cite{yang2022remix} & WSI & \ding{51}  & bag prototype & mask \& mix & \ding{55} & --- \\
RankMix (2023) \cite{Chen2023rankmix} & WSI & \ding{51}  & ranking and dropping & interpolate & \ding{51} & bag interpolation scale \\
\textbf{PseMix} (\textbf{ours}) & WSI & \ding{51}  & pseudo-bag &  mask \& r-mix & \ding{51} & pseudo-bag mixing ratio \\
\bottomrule
\end{tabular}
\end{table*}

\subsection{Multiple instance learning for WSI classification}

Multiple instance learning (MIL) \cite{ilse2018attn} nowadays has been increasingly used in computational pathology for weakly-supervised WSI classification \cite{Mercan8055584,Campanella2019,shao2021weakly,Lu2021,li2021dual,shao2021transmil,Wang9765508,zhang2022dtfd,liu2023advmil,Zhu2023,Zheng2023,Shi10119190}. Its regular procedure \cite{Chen2022} on gigapixel WSIs could be decomposed into two main stages: i) WSI preprocessing, each digital slide is transformed into a bag of instances through patching and patch feature extracting; and ii) weakly-supervised MIL, a bag is compressed into a global vector for slide-level prediction by a MIL network \cite{ilse2018attn}. The first stage is computationally intensive because of the transformation from gigapixel images to massive feature vectors. Thus, this stage is usually fixed for saving computational costs, not involved in regular training \cite{zaffar2022embedding,Kang_2023_CVPR}. 

\subsection{Mixup data augmentation}

The basic idea of Mixup \cite{zhang2018mixup} could be formulated as follows: 
\begin{equation}
\begin{aligned}
\widetilde{x} = \lambda x_i + (1-\lambda)x_j,\ 
\widetilde{y} = \lambda y_i + (1-\lambda)y_j,
\label{eq1}
\end{aligned}
\end{equation}
where the two samples $(x_i, y_i)$ and $(x_j, y_j)$ are drawn from training data, and $\lambda\in[0,1]$ follows a Beta distribution with a parameter $\alpha\in[0,+\infty]$. Despite its simplicity, Mixup has been proven, empirically and theoretically, to be an effective and practical data augmentation strategy across many different fields for improving the generalization and robustness of deep learning models \cite{yun9008296,pmlr-v119-kim20b,chen9879590,pmlr-v162-han22c,zhang2021how,guo9305717,wang2023pointpatchmix,Gong9802503}. 

From an alignment perspective, we could observe the two critical factors in the Mixup formulation given by Equation \ref{eq1}. (1) Size alignment, \textit{i.e.}, $x_i$ and $x_j$ must be aligned in size at every dimension involving mixing (\textit{e.g.}, height and width dimensions for two input gray images). (2) Semantic alignment, \textit{i.e.}, the mixed input $\widetilde{x}$ is derived from $x_i$ and $x_j$, and its content is controlled by $\lambda$, so its corresponding label $\widetilde{y}$ should be associated with $y_i$ and $y_j$, and its value should be determined by the same $\lambda$. Based on this perspective, the state-of-the-art Mixup variants for different data sources could be summarized in a unified form, as exhibited in the top half of Table \ref{tb0}. 

Apart from interpolation, masking is also one of the most important means for input mixing implementation. In the domain of image classification, it is reported \cite{chen9879590} that masking-based Mixup variants often perform better than interpolation-based ones because interpolation operation treats background and foreground equally and often yield meaningless samples. 
However, these leading Mixup methods could not be directly or efficiently applied to gigapixel WSIs, as WSIs are yet another kind of heterogeneous data, usually cast as irregular bags containing many high-dimensional instances \cite{Lu2021,Chen2022}. 

\subsection{Data augmentation for multiple instance learning}

Traditional image-level data augmentation methods \cite{shorten2019survey}, such as flipping, rotation, blurring, etc, have been adopted in WSI analysis \cite{Tellez2019,Cheng2021robust}. They are utilized in the stage of WSI preprocessing to generate different instance features for the same patch image, aiming to increase the diversity of instance-level features. However, this manner is computationally expensive to be adopted in MIL training, because there are usually tens of thousands of patch images and each one needs multiple feature extraction for a single WSI \cite{zaffar2022embedding,Kang_2023_CVPR}. 

The other methods, specially designed for MIL-based WSI classification, could be roughly divided into two categories: i) embedding-level data augmentation and ii) instance-level or bag-level data augmentation. The former focuses on augmenting instance embeddings with bag prototypes \cite{yang2022remix}, generative adversarial networks \cite{zaffar2022embedding}, or diffusion models \cite{shao2023augdiff}. The latter mainly generates new subsets from bags via hierarchical \cite{shao2021weakly,yang2023protodiv} or random sampling \cite{zhang2022dtfd}. However, these methods have not yet explored the basic idea of Mixup. 

As shown in the bottom half of Table \ref{tb0}, although ReMix \cite{yang2022remix} proposes to mix two bag inputs, it restricts the mix within the same class, not falling into the range of Mixup \cite{zhang2018mixup}. Most recently, RankMix \cite{Chen2023rankmix}, as a Mixup variant, studies the basic idea of Mixup for WSI classification. It strictly follows the original interpolation way for bag mixing. However, it heavily relies on MIL networks to obtain instance attention scores for importance ranking and instance alignment. Unfortunately, there are often potential biases between attention scores and instance importance, as highlighted in \cite{cui2023bayesmil}. Accordingly, there is still room for exploring Mixup on WSI data. 

\begin{figure*}[htbp]
\centering
\includegraphics[width=2\columnwidth]{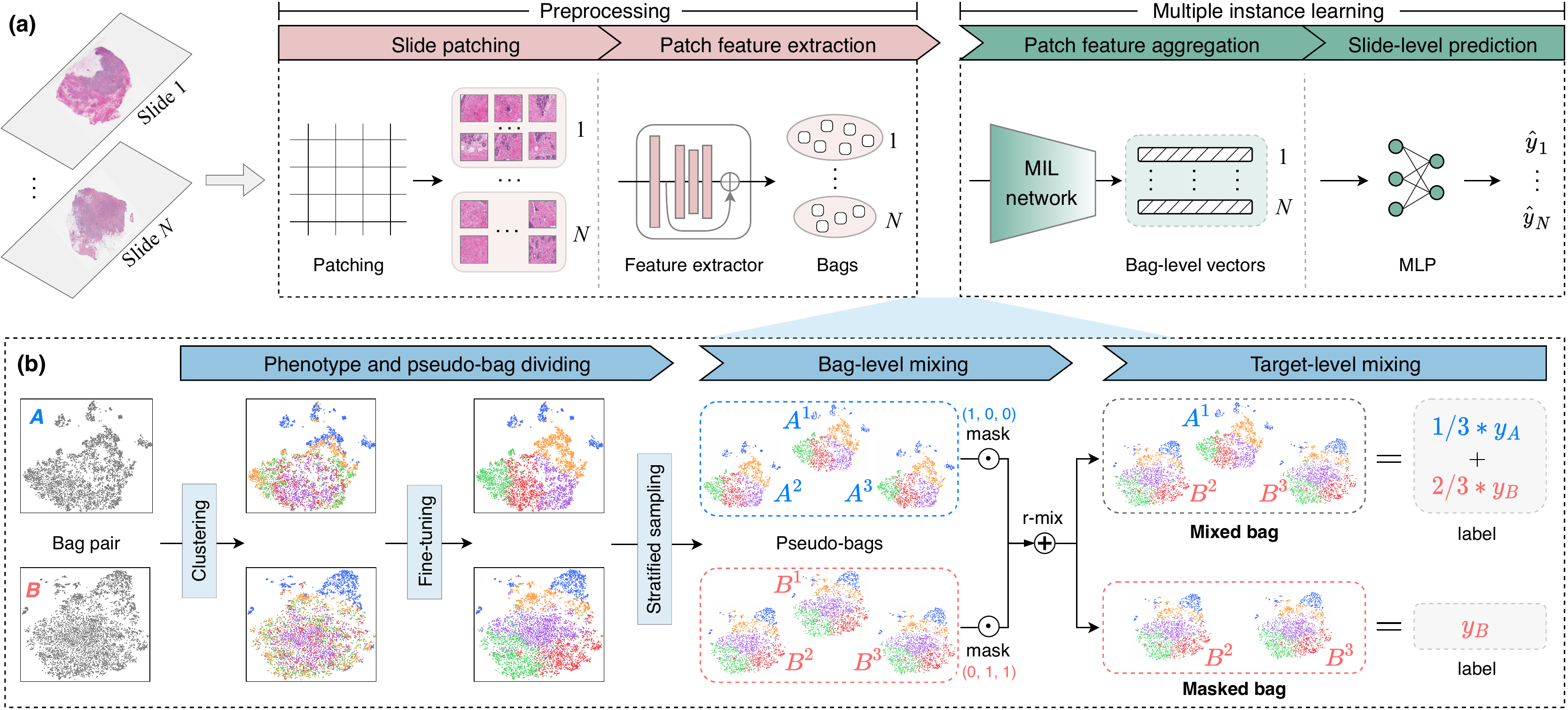}
\caption{PseMix data augmentation for MIL-based WSI classification. (a) Classical MIL paradigm for weakly-supervised WSI analysis. (b) Illustration of PseMix. Two WSI bags $A$ and $B$ are taken as examples. Solid rectangular boxes give the t-SNE visualization of instances, where scatter points represents instances and each color indicates a specific phenotype. $A^1,A^2,...,B^3$ are pseudo-bag notations. R-mix means random mixing.}
\label{fig:intro-overview}
\end{figure*}

\section{Method}
\label{sec:met}

In this section, we elaborate on how Mixup can be generalized to WSI classification via pseudo-bags, and introduce our random mixing mechanism. In the design of PseMix, we consider algorithm efficiency to make it practical and efficient to apply. Moreover, our PseMix is decoupled from the stage of MIL, so it is a plugin-and-play method for MIL models. 

\subsection{Preliminary}
\subsubsection{Notation and convention}

As shown in Figure \ref{fig:intro-overview}(a), a single digital slide is sliced into image patches, and then is preprocessed into high-dimensional vectors by a parameter-fixed feature extractor. These preprocessed feature vectors are taken as a bag of multiple instances. Given $N$ slides, we denote their processed data by 
\begin{equation}
\mathcal{D}=\{(X_i\in\mathbb{R}^{m_i\times d},y_i)\}_{i=1}^{N},
\label{eq2}
\end{equation}
where $X_i$ is the $i$-th bag, $m_i$ is instance number, $d$ means the dimension of each feature vector, and $y_i$ is a slide-level label. Let $x_{i,j}\in\mathbb{R}^{d}$ denote the $j$-th instance of the $i$-th bag, hence $X_i=\{x_{i,j}\}_{j=1}^{m_i}$. Let $\{X_i^{\tau}\}_{\tau=1}^{n}$ represent the $n$ pseudo-bags of $X_i$, where $\tau$ denotes the index of pseudo-bag. 

\subsubsection{Multiple instance learning}

In the absence of instance-level labels, most MIL framework usually takes WSI bags as input and leverages a neural network to learn bag-level representations by aggregating multiple instance embeddings, followed by an MLP (multi-layer perceptron) to output slide-level predictions, $\hat{y}_1,...,\hat{y}_N$, as illustrated in Figure \ref{fig:intro-overview}(a). This framework is also called embedding-level MIL \cite{ilse2018attn}. 

State-of-the-art MIL networks contain ABMIL \cite{ilse2018attn}, DSMIL \cite{li2021dual}, and TransMIL \cite{shao2021transmil}. They are frequently adopted in weakly-supervised WSI classification. This paper will mainly employ them as MIL networks for experiments. Note that the parameters of feature extractor are fixed and not updated with MIL training. For the feature extractor, we simply follow CLAM \cite{Lu2021} to utilize a truncated ResNet-50 model \cite{He2016} pre-trained on ImageNet \cite{Deng2009}; and $d=1024$. 

\subsection{Pseudo-bag Mixup augmentation}
\label{sec:pse}

As shown in Figure \ref{fig:intro-overview}(b), our scheme has three necessary steps: i) phenotype and pseudo-bag dividing, ii) bag-level mixing, and iii) target-level mixing. Next, we will describe their implementation details one by one. 

\subsubsection{Phenotype and pseudo-bag division}
\label{sec:psediv}

To align bag inputs in size for mixing, we cluster the instances of each bag into $l$ different phenotypes, and then adopt phenotype-stratified sampling to obtain $n$ instance-disjoint pseudo-bags. Pseudo-bag division consists of two major steps. Their details are provided in Algorithm \ref{alg1}. 

(1) \textit{Phenotype clustering}. In this step, i) we first obtain $l$ \textit{initial} instance clusters, following a bag-prototype-based clustering method \cite{yang2023protodiv}; ii) then we fine-tune these clusters for $k$ times, using an iterative cluster updating algorithm similar to that used in K-means.
Specifically, for each WSI bag, its prototype is calculated using the mean of its instance features, as shown in Algorithm \ref{alg1}. Then cosine similarity is measured between each instance and bag prototype. Based on these similarity metrics, all instances are grouped into $l$ initial clusters, where the $c$-th cluster contains the instances whose similarity scores are in $[-1+\frac{2(c-1)}{l},-1+\frac{2c}{l})$. However, these initial clusters could not be desirable, as shown in Figure \ref{fig:intro-overview}(b). One possible reason is that mean-feature-based calculation only approximately estimates the prototype of WSI bag and it thus may lead to the potential bias of instance similarity measurement. Therefore, we fine-tune those initial clusters to further improve the quality of instance clustering. Each instance cluster could be interpreted as a specific pathology phenotype \cite{shao2021weakly}. 

\begin{algorithm}[tbp]
    \caption{Pseudo-bag division (for a single bag).}
    \label{alg1}
    \KwIn{a WSI bag $X_i=\{x_{i,j}\}_{j=1}^{m_i}$, pseudo-bag number $n$, phenotype number $l$, fine-tuning times $k$.}
    \KwOut{pseudo-bags $\{X_i^{\tau}\}_{\tau=1}^{n}$.} 
    
    \tcp{Prototype-based clustering}
    
    $p_i \leftarrow \tfrac{1}{m_i}\sum_{j}x_{i,j}$ \tcp{bag prototype}
    
    initialize phenotype indicator $C_i=\{c_{i,j}\}_{j=1}^{m_i}$
    
    \For{$j\leftarrow 1$ \KwTo $m_i$}{
        $s_{i,j} \leftarrow \text{cos}(p_i,x_{i,j})$ \tcp{cosine similarity}
        
        let $c \in\{1,2,\cdots,l\}$ 
        
        $c_{i,j} \leftarrow$ find a $c \text{ s.t. } s_{i,j}\in[-1+\frac{2(c-1)}{l},-1+\frac{2c}{l})$
    }
    
    \tcp{Phenotype fine-tuning}
    
    \For{$t\leftarrow 1$ \KwTo $k$}{
    
        initialize phenotype centroids $\{f_{i,c} \in \mathbb{R}^d\}_{c=1}^{l}$
    
        \For{$c\leftarrow 1$ \KwTo $l$}{
             
            $I_c \leftarrow \{j\mid  c_{i,j}=c\}$
             
            $f_{i,c} \leftarrow \frac{1}{|I_c|}\sum_{j\in I_c}x_{i,j}$ 
        }
       
        \For{$j\leftarrow 1$ \KwTo $m_i$}{
           
            $c_{i,j}\leftarrow \mathop{\arg\min}\limits_{c} \text{cos}(f_{i,c},x_{i,j})$
        
        }
    }
    
    \tcp{Phenotype-stratified sampling}
    
    initialize $n$ empty pseudo-bags $\{X_i^{\tau}=\varnothing\}_{\tau=1}^{n}$
    
    \For{$c\leftarrow 1$ \KwTo $l$}{
             
        $I_c \leftarrow \{j\mid  c_{i,j}=c\}$
             
        randomly and uniformly split $I_c$ into $n$ parts
        
        fetch the instances \textit{w.r.t.} the $n$ parts of $I_c$
        
        append the instances to $X_i^{1},...,X_i^{n}$, respectively
    }

\end{algorithm} 

We note that the attention-based prototype proposed in \cite{yang2023protodiv} may mitigate the bias of similarity measurement and obtain better results. However, it relies on the attention score produced by MIL models. This would increase computation costs and make our PseMix dependent on attention-based models and tightly coupled with MIL training. Thus, it is not used herein to calculate initial instance clusters. 
Besides, the clustering algorithm we adopt here could be taken as a simple variant of classical K-means. Compared with K-means, this variant calculates initial clusters by a different method and updates clusters by a same iterative mechanism but with a different stopping condition. We empirically find that it could often run faster and obtain better overall results than classical K-means. Please refer to Section \ref{sec:abl} for more comparisons and discussions.

(2) \textit{Phenotype-stratified sampling}. We randomly sample instances without replacement from each phenotype stratum, obtaining $n$ pseudo-bags (each with roughly the same instance numbers) from each bag. As a result, the time and space complexity of Algorithm \ref{alg1} are $\mathcal{O}(lkm_i)$ and $\mathcal{O}(m_i)$, respectively. Given that $l$ and $k$ are usually set in $[{10}^0,{10}^1]$, the procedure of pseudo-bag division can be executed in a linear time and space complexity \textit{w.r.t.} instance number for each bag. Its actual computational cost is measured through experiments and presented in Section \ref{sec:abl}. 

(3) \textit{More discussions}. Previous studies of pseudo-bags \cite{shao2021weakly,zhang2022dtfd,yang2023protodiv} assume that \textit{pseudo-bags could inherit labels from their parent bag}. This paper also follows this basic assumption. 
However, this assumption is an ideal case because the inconsistent distribution between pseudo-bags and their parent bag may incur noises to pseudo-bag labels. Accordingly, in pseudo-bag dividing, we especially fine-tune phenotype clusters and adopt the way of phenotype-stratified sampling, to i) preserve the original phenotype distribution as much as possible in those divided pseudo-bags and ii) make pseudo-bags consistent with their parent bag in semantics. These strategies could make the basic assumption of pseudo-bag more reasonable and feasible, contributing to the semantic alignment in target mixing (introduced in Section \ref{sec-target-mixing}).
We will further evaluate and discuss different pseudo-bag division methods through comparative experiments. 

% through calculating mean prototype for each bag, measuring cosine similarity between each instance and this bag prototype, and clustering instances according to cosine similarities
% we calculate a mean prototype for $X_i$ by $f_i=\sum_{j=1}^{m_i}x_{i,j}$ and measure the cosine similarity between each instance and this prototype $f_i$. Then, we obtain $l$ instance clusters (each represents one phenotype) according to the cosine similarity of each instance, namely, an instance belongs to $\iota$-th phenotype if its value of cosine similarity is in $[-1+\frac{2(\iota-1)}{l},-1+\frac{2\iota}{l}\]$.

\subsubsection{Bag-level mixing}
\label{sec-bag-mixing}

As illustrated in Figure \ref{fig:intro-overview}(b), we augment WSI bags in this step by masking and random mixing (r-mix) on two input bags. Without loss of generality, let $(A,B)$ denote any bag pair from original training samples, where $A,B\in\{1,2,...,N\}$ and $A\neq B$. We further denote their respective pseudo-bags by $\{X_A^{\tau}\}_{\tau=1}^{n}$ and $\{X_B^{\tau}\}_{\tau=1}^{n}$, where pseudo-bags could be arranged in arbitrary orders. 

(1) \textit{Masking}. We randomly mask $1-\lambda$ and $\lambda$ pseudo-bags in $A$ and $B$, respectively. Similar to Mixup \cite{zhang2018mixup}, $\lambda\in[0,1]$ is drawn from a Beta distribution, \textit{i.e.}, $\lambda \sim \text{Beta}(\alpha, \alpha)$, where $\alpha$ is non-negative. We write two masked bags as follows:
\begin{equation}
X_A^{\prime}=(1-\mathbf{M}_{\lambda})\odot \{X_A^{\tau}\}_{\tau=1}^{n},\  X_B^{\prime}=\mathbf{M}_{\lambda}\odot \{X_B^{\tau}\}_{\tau=1}^{n},
\label{eq2.1}
\end{equation}
where $\mathbf{M}_{\lambda}\in{\{0,1\}}^n$ is a binary mask indicating which pseudo-bag to mask and keep, $\sum\mathbf{M}_{\lambda}=\lfloor\lambda(n+1)\rfloor$, and $\odot$ represents element-wise product. 

(2) \textit{Random mixing}. As depicted in Figure \ref{fig:intro-overview}(b), our r-mix operation could output two kinds of augmented bags, \textit{i.e.}, mixed or masked bags. Specifically, in r-mix, we let the masked bag $X_B^{\prime}$ join with $X_A^{\prime}$ or not join but directly output. For the first case, the output is a mixed bag:
\begin{equation}
\widetilde{X}=(1-\mathbf{M}_{\lambda})\odot \{X_A^{\tau}\}_{\tau=1}^{n}\ \cup\ \mathbf{M}_{\lambda}\odot \{X_B^{\tau}\}_{\tau=1}^{n}.
\label{eq3}
\end{equation}
It also contains $n$ pseudo-bags. In Equation \ref{eq3}, both of $A$ and $B$ are cast as $n$ pseudo-bags and pseudo-bags are the minimum unit in mixing. All of these are the necessary prerequisites for size alignment and subsequent complementary masking operations (\textit{i.e.}, $\mathbf{M}_{\lambda}$ and $1-\mathbf{M}_{\lambda}$). For the second case, \textit{i.e.}, not joining, the output is a masked bag: 
\begin{equation}
X^{\prime}=\mathbf{M}_{\lambda}\odot \{X_B^{\tau}\}_{\tau=1}^{n}.
\label{eq3.1}
\end{equation}
It is utilized as the other kind of augmented bags for training. 

(3) \textit{Motivation behind random mixing}. We introduce a simple random mixing mechanism into standard Mixup for two purposes: \textbf{i)} enhancing the diversity of training samples and \textbf{ii)} making models efficiently learn from vicinity samples (mixed bags). Pseudo-bag-masked bags could be viewed as the intermediate samples between original training data and synthetic mixed data. They could smooth the transition from original distribution to vicinity distribution, thereby helping models to efficiently learn from synthetic vicinity samples. We will evaluate and discuss our r-mix in ablation study.

From Equation \ref{eq2.1}, \ref{eq3}, and \ref{eq3.1}, we can see that our bag-level mixing is decoupled from the stage of MIL. Consequently, our scheme could serve as a plugin-and-play data augmentation method for MIL, different from the RankMix \cite{Chen2023rankmix} that relies on MIL model predictions for bag alignment. 

\subsubsection{Target-level mixing}
\label{sec-target-mixing}

For the mixed bag $\widetilde{X}$, we adopt the same $\lambda$ as that used in Equation \ref{eq3} for target mixing, \textit{i.e.}, 
\begin{equation}
\widetilde{y}=\lambda y_A + (1-\lambda) y_B.
\label{eq5}
\end{equation}
Intuitively, there are $\lambda$ and $1-\lambda$ pseudo-bags from $A$ and $B$, in $\widetilde{X}$. Given the basic assumption that pseudo-bags share the same label as their parent bag, the interpolation scale of $y_A$ and $y_B$ could be set to $\lambda$ (based on pseudo-bag mixing ratio) for semantic alignment. For the masked bag $X^{\prime}$, we set $y^{\prime}=y_B$, also following that basic assumption. 

Let $\mathcal{D}_{\text{aug}}$ denote the final training data augmented by our PseMix. It thus contains $(\widetilde{X}, \widetilde{y})$ and $(X^{\prime},y^{\prime})$. In our implementation, we adjust their proportion in $\mathcal{D}_{\text{aug}}$ by adopting a hyper-parameter $p\in[0,1]$ to set the probability of mixing in r-mix. Therefore, for any sample $\mathcal{D}_{\text{aug}}^s\in \mathcal{D}_{\text{aug}}$, it is written as follows:
\begin{equation}
\mathcal{D}_{\text{aug}}^s=
\begin{cases}
(\widetilde{X},\  \widetilde{y})& \text{with a probability of } p,\\
(X^{\prime},\  y^{\prime})& \text{with a probability of } 1-p.
\end{cases}
\label{eq7}
\end{equation}

\section{Experiments}
\label{sec:exp}

In this section, we mainly evaluate the effectiveness and advantages of PseMix through comparative experiments and ablation studies.
We describe experimental settings in Section \ref{sec:set}. Then we validate PseMix using three different WSI datasets and compare it with other mixing data augmentation methods in Section \ref{sec:clf}. In Section \ref{sec:gen} and \ref{sec:rob}, we further show the advantages of PseMix in generalization and robustness. Finally, ablation studies and hyper-parameter sensitivity analysis are presented in Section \ref{sec:abl} and \ref{sec:fur}, respectively. 

\subsection{Experimental settings}
\label{sec:set}

\subsubsection{Datasets and tasks}
We use the following pathology datasets in this study: i) TCGA-BRCA, the subtyping of Invasive Ductal Carcinoma and Invasive Lobular Carcinoma for invasive breast carcinoma, ii) TCGA-LUNG, the recognition of Lung Adenocarcinoma and Lung Squamous Cell Carcinoma in Non-Small Cell Lung Carcinoma, and iii) TCGA-RCC, the classification of Clear Cell, Papillary, and Chromophobe for Renal Cell Carcinoma. These three datasets are often adopted to evaluate MIL models \cite{shao2021transmil,Chen2022}. They are publicly-available at TCGA (\href{https://portal.gdc.cancer.gov}{https://portal.gdc.cancer.gov}). As shown in Table \ref{tb1}, a total of 2,740 slides are collected. They are preprocessed with the tools developed by CLAM \cite{Lu2021}.

\begin{table}[htbp]
\centering
\caption{Statistical details of three WSI datasets.}
\label{tb1}
\begin{tabular}{l|c|c|c}
\toprule
Item       & TCGA-BRCA & TCGA-LUNG          & TCGA-RCC          \\ \midrule
\# Patients          & 898         & 947           & 895          \\
\# Slides              & 953       & 1,044         & 937 \\    
\# Patches           & 2,961,552   & 3,235,064     & 3,317,384    \\
\# Patches per slide & 3,107.6   & 3098.7     & 3,540.4    \\ \bottomrule
\end{tabular}
\end{table}

\begin{table*}[htbp]
\centering
\caption{Classification performance of vanilla MIL models and PseMix-based ones on three WSI datasets.}
\label{tb2}
\begin{tabular}{l|l|cc|cc|cc|cc}
\toprule
\multirow{2}{*}{Network}  & \multirow{2}{*}{Method} & \multicolumn{2}{c|}{TCGA-BRCA} & \multicolumn{2}{c|}{TCGA-LUNG} & \multicolumn{2}{c|}{TCGA-RCC} & \multicolumn{2}{c}{Average (\%)} \\ %\cmidrule(lr){3-4} \cmidrule(lr){5-6} \cmidrule(lr){7-8}
                          &                         & ACC (\%)        & AUC (\%)       & ACC (\%)        & AUC (\%)        & ACC (\%)       & AUC (\%) & ACC       & AUC \\ \midrule
\rowcolor{gray!15} \cellcolor{white} \multirow{7}{*}{ABMIL \cite{ilse2018attn}}    & vanilla                 & 86.34{\tiny\ $\pm$ 1.97}  & 87.05{\tiny\ $\pm$ 2.97} & 83.88{\tiny\ $\pm$ 4.94} & 92.23{\tiny\ $\pm$ 2.81} & 88.90{\tiny\ $\pm$ 1.20} & 97.36{\tiny\ $\pm$ 1.10}  & 86.37 & 92.21          \\
%  & w/ Pseudo-bag               & 85.71{\tiny\ $\pm$ 2.25} & 87.39{\tiny\ $\pm$ 2.38} & 85.11{\tiny\ $\pm$ 2.60} & 92.19{\tiny\ $\pm$ 1.87} & 90.38{\tiny\ $\pm$ 0.84} & 97.94{\tiny\ $\pm$ 0.74} & 87.07 & 92.51     \\
  & w/ ReMix               & 86.46{\tiny\ $\pm$ 1.67} & 87.74{\tiny\ $\pm$ 3.97} & 84.48{\tiny\ $\pm$ 3.99} & 91.66{\tiny\ $\pm$ 2.04} & 88.24{\tiny\ $\pm$ 1.99} & 96.93{\tiny\ $\pm$ 0.90}  & 86.39 & 92.11 \\
  & w/ Mixup               & 85.91{\tiny\ $\pm$ 2.09} & 87.31{\tiny\ $\pm$ 3.54} & 86.02{\tiny\ $\pm$ 2.67} & 92.75{\tiny\ $\pm$ 2.47} & 90.39{\tiny\ $\pm$ 2.05} & 97.78{\tiny\ $\pm$ 0.76}  & 87.44 & 92.61  \\
  & w/ RankMix              & 85.62{\tiny\ $\pm$ 1.34} & 87.48{\tiny\ $\pm$ 3.97} & 84.73{\tiny\ $\pm$ 1.70} & 91.76{\tiny\ $\pm$ 2.07} & 89.44{\tiny\ $\pm$ 2.00} & 97.37{\tiny\ $\pm$ 0.68} & 86.60 & 92.20  \\
  & w/ InstanceMix               & 86.24{\tiny\ $\pm$ 1.94} & 87.98{\tiny\ $\pm$ 3.18} & 84.83{\tiny\ $\pm$ 2.76} & 91.74{\tiny\ $\pm$ 2.22} & 88.73{\tiny\ $\pm$ 1.96} & 97.60{\tiny\ $\pm$ 0.41} & 86.60 & 92.44  \\
\rowcolor{yellow!15} \cellcolor{white}  & w/ \textbf{PseMix} (ours) & \textbf{86.64}{\tiny\ $\pm$ 3.11} & \textbf{89.49}{\tiny\ $\pm$ 3.69} & \textbf{86.45}{\tiny\ $\pm$ 3.28} & \textbf{93.01}{\tiny\ $\pm$ 2.06} & \textbf{90.50}{\tiny\ $\pm$ 1.66} & \textbf{98.02}{\tiny\ $\pm$ 0.55}   & \textbf{87.86} & \textbf{93.51} \\ 
  & $\Delta$ over vanilla              & \textcolor[RGB]{36,141,88}{+ 0.30} & \textcolor[RGB]{36,141,88}{+ 2.44} & \textcolor[RGB]{36,141,88}{+ 2.57} & \textcolor[RGB]{36,141,88}{+ 0.78} & \textcolor[RGB]{36,141,88}{+ 1.60} & \textcolor[RGB]{36,141,88}{+ 0.66} & \textcolor[RGB]{36,141,88}{+ 1.49} & \textcolor[RGB]{36,141,88}{+ 1.30}         \\ 
                         \midrule  \midrule
\rowcolor{gray!15} \cellcolor{white} \multirow{7}{*}{DSMIL\cite{li2021dual}}    & vanilla                 & 86.75{\tiny\ $\pm$ 1.48} & 87.73{\tiny\ $\pm$ 2.04} & 85.70{\tiny\ $\pm$ 3.18} & 92.99{\tiny\ $\pm$ 2.94} & 89.81{\tiny\ $\pm$ 3.29} & 97.65{\tiny\ $\pm$ 0.81} & 87.42 & 92.79          \\
%  & w/ Pseudo-bag               & 85.72{\tiny\ $\pm$ 1.63} & 89.31{\tiny\ $\pm$ 2.83} & 86.96{\tiny\ $\pm$ 2.91} & 93.57{\tiny\ $\pm$ 2.49} & 90.50{\tiny\ $\pm$ 1.79} & \textbf{97.93}{\tiny\ $\pm$ 0.48} & 87.73 & 93.60    \\
  & w/ ReMix               & 85.74{\tiny\ $\pm$ 2.38} & 87.98{\tiny\ $\pm$ 3.59} & 83.30{\tiny\ $\pm$ 3.71} & 91.59{\tiny\ $\pm$ 2.06} & 87.49{\tiny\ $\pm$ 1.30} & 96.54{\tiny\ $\pm$ 1.74}  & 85.51 & 92.04  \\
  & w/ Mixup               & 86.35{\tiny\ $\pm$ 2.26} & 88.30{\tiny\ $\pm$ 3.10} & 87.52{\tiny\ $\pm$ 3.12} & \textbf{94.22}{\tiny\ $\pm$ 2.39} & 90.60{\tiny\ $\pm$ 2.29} & 97.77{\tiny\ $\pm$ 0.85}  & 88.16 & 93.43  \\
  & w/ RankMix              & 84.80{\tiny\ $\pm$ 1.73} & 86.49{\tiny\ $\pm$ 2.58} & 86.39{\tiny\ $\pm$ 3.33} & 93.51{\tiny\ $\pm$ 2.69} & 90.27{\tiny\ $\pm$ 1.25} & 97.38{\tiny\ $\pm$ 0.87} & 87.15 & 92.46  \\
  & w/ InstanceMix               & 85.39{\tiny\ $\pm$ 3.23} & 87.94{\tiny\ $\pm$ 3.48} & 86.45{\tiny\ $\pm$ 3.39} & 92.40{\tiny\ $\pm$ 2.09} & 89.48{\tiny\ $\pm$ 2.28} & 97.54{\tiny\ $\pm$ 0.55} & 87.11 & 92.63 \\
\rowcolor{yellow!15} \cellcolor{white}   & w/ \textbf{PseMix} (ours) & \textbf{88.22}{\tiny\ $\pm$ 2.65} & \textbf{89.65}{\tiny\ $\pm$ 3.19} & \textbf{88.68}{\tiny\ $\pm$ 2.19} & 93.92{\tiny\ $\pm$ 2.19} & \textbf{90.62}{\tiny\ $\pm$ 1.20} & \textbf{97.89}{\tiny\ $\pm$ 0.44}   & \textbf{89.17}  & \textbf{93.82} \\ %\cmidrule(lr){2-2}
  & $\Delta$ over vanilla            & \textcolor[RGB]{36,141,88}{+ 1.47} & \textcolor[RGB]{36,141,88}{+ 1.92} & \textcolor[RGB]{36,141,88}{+ 2.98} & \textcolor[RGB]{36,141,88}{+ 0.93} & \textcolor[RGB]{36,141,88}{+ 0.81} & \textcolor[RGB]{36,141,88}{+ 0.24} & \textcolor[RGB]{36,141,88}{+ 1.75} & \textcolor[RGB]{36,141,88}{+ 1.03}         \\  
                         \midrule  \midrule
\rowcolor{gray!15} \cellcolor{white} \multirow{7}{*}{TransMIL \cite{shao2021transmil}} & vanilla                 & 85.31{\tiny\ $\pm$ 0.65} & 88.83{\tiny\ $\pm$ 1.37} & 85.31{\tiny\ $\pm$ 4.01} & 92.14{\tiny\ $\pm$ 2.56} & 90.61{\tiny\ $\pm$ 1.62} & 97.88{\tiny\ $\pm$ 0.80}   & 87.08 & 92.95     \\
%  & w/ Pseudo-bag          & 86.56{\tiny\ $\pm$ 0.91} & 89.31{\tiny\ $\pm$ 1.22} & 87.13{\tiny\ $\pm$ 3.37} & 93.33{\tiny\ $\pm$ 1.61} & 90.16{\tiny\ $\pm$ 1.82} & 97.69{\tiny\ $\pm$ 0.94} & 87.95 & 93.44     \\
  & w/ ReMix               & 79.88{\tiny\ $\pm$ 2.12} & 78.76{\tiny\ $\pm$ 4.17} & 83.27{\tiny\ $\pm$ 3.05} & 90.67{\tiny\ $\pm$ 3.02} & 87.52{\tiny\ $\pm$ 2.53} & 96.60{\tiny\ $\pm$ 1.04}  & 83.56 & 88.68  \\
  & w/ Mixup               & 81.52{\tiny\ $\pm$ 2.39} & 88.41{\tiny\ $\pm$ 3.07} & 87.62{\tiny\ $\pm$ 3.76} & \textbf{94.52}{\tiny\ $\pm$ 2.49} & 91.03{\tiny\ $\pm$ 1.60} & 97.88{\tiny\ $\pm$ 0.64} & 86.72 & 93.60 \\
  & w/ RankMix              & 84.69{\tiny\ $\pm$ 1.13} & 87.73{\tiny\ $\pm$ 1.31} & 86.45{\tiny\ $\pm$ 2.66} & 93.75{\tiny\ $\pm$ 2.54} & 90.63{\tiny\ $\pm$ 1.92} & \textbf{97.97}{\tiny\ $\pm$ 0.51} & 87.26 & 93.15 \\
  & w/ InstanceMix               & 86.77{\tiny\ $\pm$ 2.35} & 89.44{\tiny\ $\pm$ 3.01} & 86.74{\tiny\ $\pm$ 2.25} & 92.72{\tiny\ $\pm$ 2.34} & 91.04{\tiny\ $\pm$ 1.57} & 97.84{\tiny\ $\pm$ 0.31} & 88.18 & 93.33 \\
\rowcolor{yellow!15} \cellcolor{white}   & w/ \textbf{PseMix} (ours)  & \textbf{86.98}{\tiny\ $\pm$ 1.47} & \textbf{90.40}{\tiny\ $\pm$ 2.29} & \textbf{87.67}{\tiny\ $\pm$ 2.80} & 93.47{\tiny\ $\pm$ 1.91} & \textbf{91.14}{\tiny\ $\pm$ 1.94} & 97.76{\tiny\ $\pm$ 0.69} & \textbf{88.60} & \textbf{93.88}  \\ %\cmidrule(lr){2-2}
  & $\Delta$ over vanilla              & \textcolor[RGB]{36,141,88}{+ 1.67} & \textcolor[RGB]{36,141,88}{+ 1.57} & \textcolor[RGB]{36,141,88}{+ 2.36} & \textcolor[RGB]{36,141,88}{+ 1.33} & \textcolor[RGB]{36,141,88}{+ 0.53} & \textcolor[RGB]{128,128,128}{-- 0.12} & \textcolor[RGB]{36,141,88}{+ 1.52} & \textcolor[RGB]{36,141,88}{+ 0.93}          \\ 
\bottomrule
\end{tabular}
\end{table*}

\subsubsection{Performance evaluation}

Due to the relatively small size of WSI datasets, we follow the study design of other MIL-based WSI classification works by using 4-fold cross-validation (cv) to evaluate model performance more accurately and reliably. In each of 4 training sessions, a validation set is randomly split from training data for early stopping and model selection, resulting in a ratio of 65:10:25 for the size of training, validation, and test sets on the patient level. 
The two classification metrics, Area Under the Curve (AUC) and accuracy (ACC), are used. And we report their mean and standard deviation across 4 folds. For ACC calculation, the class with the largest probability is taken as classification prediction. In addition, we further assess the generalization ability of models by measuring the AUC and loss gap between training and test set, following \cite{jiang2018predicting,zhang2021how}. 

\subsubsection{Implementation details}
%ABMIL \cite{ilse2018attn}, DSMIL \cite{li2021dual}, and TransMIL \cite{shao2021transmil} are employed as MIL backbones for experiments, as they are frequently used in WSI classification. We follow their official codes for model implementation. A truncated ResNet-50 model \cite{He2016} pre-trained on ImageNet \cite{Deng2009} is utilized as the feature extractor and $d=1024$, following \cite{Lu2021}. 
We set $\alpha$ to 1 by default, which means that $\lambda$ is distributed uniformly over $[0,1]$. For the hyper-parameters of pseudo-bag division, we empirically set $n=30$ and $l=k=8$ by default across all used datasets and networks. We will analyze important hyper-parameters in Section \ref{sec:fur}. All experiments are run on a machine with two NVIDIA GeForce RTX 3090 GPUs. Please refer to our codes (\href{https://github.com/liupei101/PseMix}{https://github.com/liupei101/PseMix}) for more details. % could be found at our public-available codes\footnote{https://github.com/liupei101/PseMix}. 

\subsection{Basic classification performance}
\label{sec:clf}

We first verify whether PseMix could help improve the classification performance of MIL models. The models trained without any data augmentation, \textit{i.e.}, vanilla models, are taken as the baseline for comparison. Moreover, we run other input mixing methods (see Table \ref{tb0}) for further comparisons:
\begin{itemize}
    \item \textit{ReMix} \cite{yang2022remix}, which mixes the prototypes of two bags but only processes the bags within the same class;
    \item \textit{Mixup} \cite{zhang2018mixup}, the original interpolation-based Mixup, in which two bags are aligned in instance number before interpolation by random dropping instances from the bag with larger instance number;
    \item \textit{RankMix} \cite{Chen2023rankmix}, an improved interpolation-based one, in which the instances of each bag are first ranked in order according to their attention scores, and then two bags are aligned in number by dropping the instances with lower score from the bag with larger instance number;
    \item \textit{InstanceMix}, an instance-level Mixup baseline, in which each bag is not divided into pseudo-bags and two input bags are mixed after randomly masking some instances, rather than pseudo-bags, according to the $\lambda$ with Beta distribution, specially designed for the comparison to our pseudo-bag-level scheme. 
\end{itemize}
%They are i) \textit{ReMix} \cite{yang2022remix}, which only mixes bags within the same class; ii) \textit{Mixup} \cite{zhang2018mixup}, the original interpolation-based Mixup in which two bags are aligned in instance number by random dropping; iii) \textit{RankMix} \cite{Chen2023rankmix}, an improved interpolation-based one in which two bags are aligned firstly by instance ranking and then by instance dropping; iv) \textit{InstanceMix}, an instance-level Mixup baseline in which two bags are mixed after masking some instances, rather than pseudo-bags, specially designed for the comparison to our pseudo-bag-level scheme. 
Besides data augmentation, other regular training settings are shared for all baselines and networks. 

There are two findings in Table \ref{tb2}. (1) PseMix almost always helps improve the performance of MIL models in WSI classification. Its average improvements over vanilla models are consistently positive across three MIL networks, ranging from 0.93\% to 1.75\%. (2) PseMix could often surpass other related input mixing methods in average ACC and AUC. Moreover, the maximum gap in average ACC between PseMix and others is 1.01\% (when it is on DSMIL); as in average AUC, the maximum is 0.90\% (when it is on ABMIL). These results suggest that our pseudo-bag Mixup scheme is an effective and superior Mixup variant for improving the performance of MIL models in WSI classification. 

\begin{figure*}[tp]
\centering
\includegraphics[width=\textwidth]{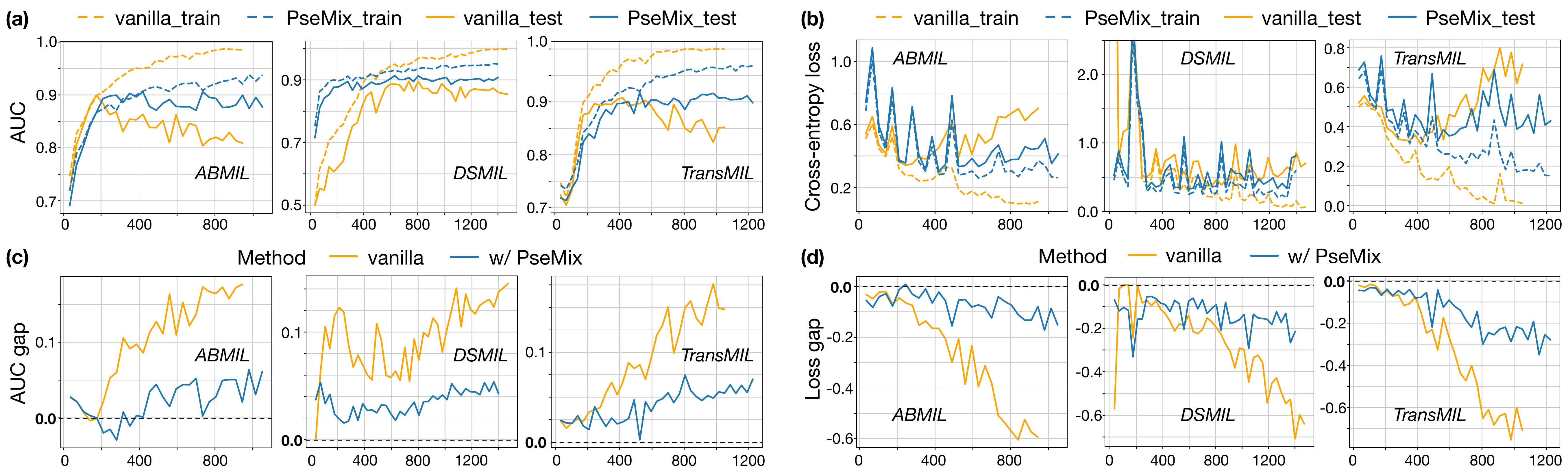}
\caption{Generalization gap (AUC and cross-entropy loss) between training and test set. Three MIL networks are trained on TCGA-BRCA.}
\label{fig:test-gene}
\end{figure*}

As described in Section \ref{sec:met}, our PseMix is a masking-based mixing, pseudo-bag-level Mixup scheme. It outperforms two interpolation-based variants and a masking-based instance-level variant in most cases. Moreover, it is superior to them in average metrics for three MIL networks. (1) For two interpolation-based variants, i.e., original Mixup and its improved RankMix, both of them often need to drop instances from the bag with more instances, to align two irregular bags for interpolation. This operation may result in artificial information loss, especially when there is a big gap in instance numbers. Furthermore, interpolation-based Mixup often performs worse than masking-based ones in image classification, since interpolation operations equally treat the background and foreground in images, as demonstrated in \cite{chen9879590}. By contrast, our PseMix does not rely on interpolation but masks pseudo-bags for bag mixing. (2) For InstanceMix, it is based on masking like PseMix but operates at the instance level. And it treats all instances equally, without the critical semantic alignment for Mixup. Our PseMix tackles this by manipulating pseudo-bags. For those reasons, PseMix could often perform better than previous Mixup variants in WSI classification. 

\subsection{Special generalization tests}
\label{sec:gen}

Most previous works of WSI classification usually focus on promoting classification performance, rarely concerned with more aspects of model generalization. Here we demonstrate more advantages of PseMix in generalization by using generalization gap metrics and out-of-distribution data.
% another benefit of PseMix, \textit{i.e.}, better model generalization.

\subsubsection{Generalization gap}

Following \cite{jiang2018predicting,zhang2021how}, we measure the generalization gap between training and test sets to show the other metrics of generalization performance. We have the following observations from Figure \ref{fig:test-gene}. (1) Vanilla models always perform better on training set but worse on test set in training. (2) By contrast, their PseMix-based counterparts often have lower test losses and higher test AUCs, showing smaller generalization gaps.
An intuitive, possible explanation is that PseMix enables MIL models to learn from the new training samples drawn from vicinity distribution and these new vicinity samples could help models to expand generalization boundary, leading to the better performance of PseMix-based models on unseen data. 

\begin{figure}[htp]
\centering
\includegraphics[width=\columnwidth]{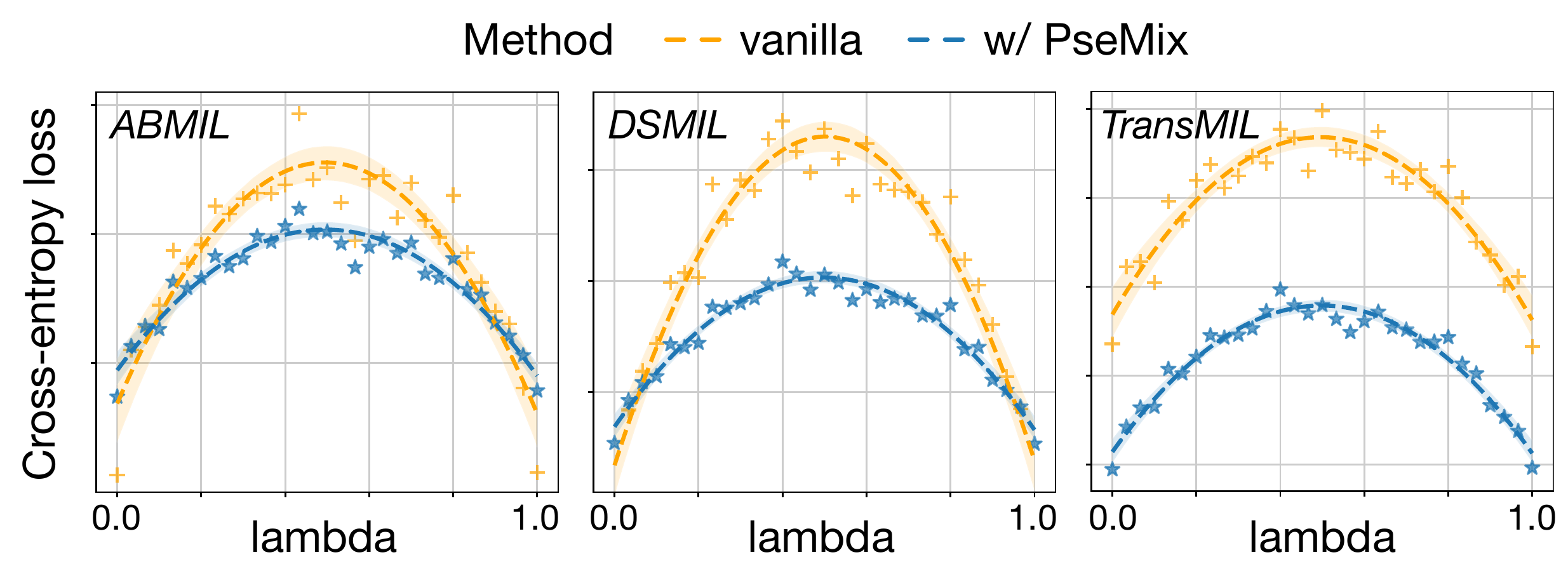}
\caption{Cross-entropy loss of in-between training data and their corresponding soft-labels.}
\label{fig:test-inbet}
\end{figure}

\begin{table*}[htbp]
\centering
\caption{Robustness against patch occlusion. }
\label{tb3}
\begin{tabular}{c|l|cc|cc|cc|cc}
\toprule
\multirow{2}{*}{Patch masking}    & \multirow{2}{*}{Method} & \multicolumn{2}{c|}{ABMIL} & \multicolumn{2}{c|}{DSMIL} & \multicolumn{2}{c|}{TransMIL} & \multicolumn{2}{c}{Average (\%)} \\ %\cmidrule(lr){3-4} \cmidrule(lr){5-6} \cmidrule(lr){7-8}
                          &                         & ACC (\%)        & AUC (\%)       & ACC (\%)        & AUC (\%)        & ACC (\%)       & AUC (\%) & ACC       & AUC       \\ \midrule
\multirow{2}{*}{20\%}    & vanilla                 & 85.60{\tiny\ $\pm$ 2.14} & 87.13{\tiny\ $\pm$ 2.65} & 85.11{\tiny\ $\pm$ 1.87} & 86.94{\tiny\ $\pm$ 2.99} & 81.93{\tiny\ $\pm$ 2.44} & 87.51{\tiny\ $\pm$ 1.49} & 84.21 & 87.19  \\
  & w/ PseMix               & \textbf{87.27}{\tiny\ $\pm$ 3.01} & \textbf{89.91}{\tiny\ $\pm$ 3.05} & \textbf{86.96}{\tiny\ $\pm$ 2.32} & \textbf{89.12}{\tiny\ $\pm$ 3.90} & \textbf{87.07}{\tiny\ $\pm$ 1.78} & \textbf{90.25}{\tiny\ $\pm$ 2.53} & \textbf{87.10} & \textbf{89.76}  \\ \midrule
                         % & $\Delta$               & \textcolor[RGB]{36,141,88}{+ 1.27} & \textcolor[RGB]{36,141,88}{+ 2.77} & \textcolor[RGB]{36,141,88}{+ 2.07} & \textcolor[RGB]{36,141,88}{+ 1.95} & \textcolor[RGB]{36,141,88}{+ 4.31} & \textcolor[RGB]{36,141,88}{+ 2.29}            \\ \midrule
\multirow{2}{*}{40\%}    & vanilla                 & 84.65{\tiny\ $\pm$ 3.02} & 86.86{\tiny\ $\pm$ 2.56} & \textbf{85.95}{\tiny\ $\pm$ 2.03} & 87.07{\tiny\ $\pm$ 3.41} & 80.57{\tiny\ $\pm$ 4.20} & 86.28{\tiny\ $\pm$ 1.98} & 83.72 & 86.74  \\
  & w/ PseMix               & \textbf{86.65}{\tiny\ $\pm$ 2.63} & \textbf{89.87}{\tiny\ $\pm$ 2.99} & 85.50{\tiny\ $\pm$ 2.77} & \textbf{88.77}{\tiny\ $\pm$ 3.94} & \textbf{84.77}{\tiny\ $\pm$ 3.63} & \textbf{89.61}{\tiny\ $\pm$ 2.02} & \textbf{85.64} & \textbf{89.42} \\ \midrule 
                         % & $\Delta$               & \textcolor[RGB]{36,141,88}{+ 1.48} & \textcolor[RGB]{36,141,88}{+ 2.90} & \textcolor[RGB]{36,141,88}{+ 0.59} & \textcolor[RGB]{36,141,88}{+ 1.36} & \textcolor[RGB]{36,141,88}{+ 2.84} & \textcolor[RGB]{36,141,88}{+ 2.82}            \\ \midrule
\multirow{2}{*}{60\%} & vanilla                 & 85.17{\tiny\ $\pm$ 3.44} & 87.05{\tiny\ $\pm$ 2.57} & 84.57{\tiny\ $\pm$ 2.25} & 86.94{\tiny\ $\pm$ 3.65} & 76.43{\tiny\ $\pm$ 10.16} & 84.26{\tiny\ $\pm$ 2.85} & 82.06 & 86.08  \\
  & w/ PseMix               & \textbf{86.65}{\tiny\ $\pm$ 2.63} & \textbf{89.64}{\tiny\ $\pm$ 3.04} & \textbf{84.77}{\tiny\ $\pm$ 2.58} & \textbf{88.85}{\tiny\ $\pm$ 4.07} & \textbf{82.27}{\tiny\ $\pm$ 6.11} & \textbf{88.02}{\tiny\ $\pm$ 2.27} & \textbf{84.56} & \textbf{88.84}  \\ \midrule
                         % & $\Delta$               & \textcolor[RGB]{36,141,88}{+ 0.43} & \textcolor[RGB]{36,141,88}{+ 2.08} & \textcolor[RGB]{36,141,88}{+ 2.61} & \textcolor[RGB]{36,141,88}{+ 1.56} & \textcolor[RGB]{36,141,88}{+ 5.63} & \textcolor[RGB]{36,141,88}{+ 3.40}            \\ \midrule
\multirow{2}{*}{80\%} & vanilla                 & 84.44{\tiny\ $\pm$ 2.85} & 87.05{\tiny\ $\pm$ 1.88} & \textbf{84.26}{\tiny\ $\pm$ 2.51} & 86.43{\tiny\ $\pm$ 3.27} & 64.44{\tiny\ $\pm$ 19.91} & 76.82{\tiny\ $\pm$ 4.90} & 77.71 & 83.43 \\
  & w/ PseMix               & \textbf{85.06}{\tiny\ $\pm$ 5.09} & \textbf{88.88}{\tiny\ $\pm$ 3.43} & 80.71{\tiny\ $\pm$ 4.31} & \textbf{87.28}{\tiny\ $\pm$ 4.71} & \textbf{78.92}{\tiny\ $\pm$ 5.80} & \textbf{84.43}{\tiny\ $\pm$ 6.18} & \textbf{81.56} & \textbf{86.86} \\ 
                         % & $\Delta$               & \textcolor[RGB]{128,128,128}{-- 1.49} & \textcolor[RGB]{36,141,88}{+ 0.60} & \textcolor[RGB]{36,141,88}{+ 1.44} & \textcolor[RGB]{36,141,88}{+ 1.59} & \textcolor[RGB]{36,141,88}{+ 13.22} & \textcolor[RGB]{36,141,88}{+ 6.18}            \\ 
\bottomrule
\end{tabular}
\end{table*}

\begin{table*}[htbp]
\centering
\caption{Robustness against label corruption.}
\label{tb4}
\begin{tabular}{c|c|l|cc|cc|cc|cc}
\toprule
\multirow{2}{*}{Checkpoint} & \multirow{2}{*}{\makecell[c]{Label\\ corruption}}    & \multirow{2}{*}{Method} & \multicolumn{2}{c|}{ABMIL} & \multicolumn{2}{c|}{DSMIL} & \multicolumn{2}{c|}{TransMIL} & \multicolumn{2}{c}{Average (\%)}  \\ %\cmidrule(lr){4-5} \cmidrule(lr){6-7} \cmidrule(lr){8-9}
                    &      &                         & ACC (\%)        & AUC (\%)       & ACC (\%)        & AUC (\%)        & ACC (\%)       & AUC (\%) & ACC      & AUC        \\ \midrule
\multirow{7}{*}{\makecell[c]{Best\\ epoch}} & \multirow{2}{*}{20\%}    & vanilla                 & \textbf{84.55}{\tiny\ $\pm$ 2.84} & \textbf{83.81}{\tiny\ $\pm$ 4.83} & 83.51{\tiny\ $\pm$ 2.00} & 83.84{\tiny\ $\pm$ 4.20} & 81.96{\tiny\ $\pm$ 1.16} & 82.32{\tiny\ $\pm$ 2.74} & 83.34 & 83.32 \\
&                          & w/ PseMix               & 84.54{\tiny\ $\pm$ 3.25} & 83.67{\tiny\ $\pm$ 4.66} & \textbf{85.80}{\tiny\ $\pm$ 3.55} & \textbf{85.27}{\tiny\ $\pm$ 4.93} & \textbf{84.87}{\tiny\ $\pm$ 1.91} & \textbf{86.41}{\tiny\ $\pm$ 3.98} & \textbf{85.07} & \textbf{85.12} \\  \cmidrule{2-11}
& \multirow{2}{*}{50\%}    & vanilla                 & 80.82{\tiny\ $\pm$ 1.77} & 76.77{\tiny\ $\pm$ 4.15} & 79.96{\tiny\ $\pm$ 1.57} & 75.71{\tiny\ $\pm$ 3.95} & 79.85{\tiny\ $\pm$ 0.15} & 74.54{\tiny\ $\pm$ 5.60} & 80.21 & 75.67  \\
&                          & w/ PseMix               & \textbf{81.35}{\tiny\ $\pm$ 2.75} & \textbf{78.72}{\tiny\ $\pm$ 2.33} & \textbf{83.62}{\tiny\ $\pm$ 1.96} & \textbf{79.94}{\tiny\ $\pm$ 4.89} & \textbf{81.95}{\tiny\ $\pm$ 0.46} & \textbf{81.61}{\tiny\ $\pm$ 2.75} & \textbf{82.31} & \textbf{80.09} \\ \cmidrule{2-11}
& \multirow{2}{*}{80\%} & vanilla                 & \textbf{78.28}{\tiny\ $\pm$ 0.99} & \textbf{62.81}{\tiny\ $\pm$ 7.83} & 79.02{\tiny\ $\pm$ 1.10} & 59.40{\tiny\ $\pm$ 2.86} & 74.99{\tiny\ $\pm$ 3.98} & 53.92{\tiny\ $\pm$ 4.25} & 77.43 & 58.71 \\
&                          & w/ PseMix               & \textbf{78.28}{\tiny\ $\pm$ 0.83} & 61.13{\tiny\ $\pm$ 8.78} & \textbf{79.75}{\tiny\ $\pm$ 1.19} & \textbf{64.81}{\tiny\ $\pm$ 2.77} & \textbf{77.53}{\tiny\ $\pm$ 2.89} & \textbf{59.76}{\tiny\ $\pm$ 6.02} & \textbf{78.52} & \textbf{61.90}  \\ \midrule
\multirow{7}{*}{\makecell[c]{Last\\ epoch}} & \multirow{2}{*}{20\%} & vanilla                 & 77.61{\tiny\ $\pm$ 4.23} & 77.07{\tiny\ $\pm$ 5.42} & 82.05{\tiny\ $\pm$ 1.17} & 80.91{\tiny\ $\pm$ 4.12} & 76.58{\tiny\ $\pm$ 2.20} & 77.31{\tiny\ $\pm$ 2.68} & 78.75 & 78.43  \\
&                          & w/ PseMix               & \textbf{81.20}{\tiny\ $\pm$ 2.29} & \textbf{80.53}{\tiny\ $\pm$ 7.14} & \textbf{83.74}{\tiny\ $\pm$ 4.66} & \textbf{82.71}{\tiny\ $\pm$ 7.41} & \textbf{82.05}{\tiny\ $\pm$ 1.31} & \textbf{82.52}{\tiny\ $\pm$ 5.29} & \textbf{82.33} & \textbf{81.92} \\ \cmidrule{2-11}
& \multirow{2}{*}{50\%} & vanilla                 & 75.34{\tiny\ $\pm$ 3.05} & 70.01{\tiny\ $\pm$ 5.81} & 75.43{\tiny\ $\pm$ 3.48} & 73.04{\tiny\ $\pm$ 4.11} & 70.35{\tiny\ $\pm$ 3.77} & 68.94{\tiny\ $\pm$ 3.37} & 73.71 & 70.66 \\
&                          & w/ PseMix               & \textbf{79.05}{\tiny\ $\pm$ 5.08} & \textbf{75.14}{\tiny\ $\pm$ 4.12} & \textbf{82.78}{\tiny\ $\pm$ 0.90} & \textbf{78.08}{\tiny\ $\pm$ 4.07} & \textbf{78.82}{\tiny\ $\pm$ 2.11} & \textbf{75.51}{\tiny\ $\pm$ 3.72} & \textbf{80.22} & \textbf{76.24}  \\ \cmidrule{2-11}
& \multirow{2}{*}{80\%} & vanilla                 & \textbf{58.03}{\tiny\ $\pm$ 10.23} & \textbf{60.34}{\tiny\ $\pm$ 5.37} & 55.75{\tiny\ $\pm$ 14.53} & 58.63{\tiny\ $\pm$ 9.27} & 53.67{\tiny\ $\pm$ 7.72} & 51.20{\tiny\ $\pm$ 3.45} & 55.82 & 56.72 \\
&                          & w/ PseMix               & 57.53{\tiny\ $\pm$ 9.73} & 60.16{\tiny\ $\pm$ 7.76} & \textbf{65.99}{\tiny\ $\pm$ 11.58} & \textbf{63.82}{\tiny\ $\pm$ 3.65} & \textbf{73.81}{\tiny\ $\pm$ 6.00} & \textbf{57.82}{\tiny\ $\pm$ 9.21} & \textbf{65.78} & \textbf{60.60} \\ 
\bottomrule
\end{tabular}
\end{table*}

\subsubsection{Test on in-between training data}

We further examine the model performance on out-of-distribution in-between data, following Mixup \cite{zhang2018mixup}. Specifically, we synthesize new in-between data for testing by applying Equation \ref{eq3} to training samples, and obtain their soft labels by the target mixing of Equation \ref{eq5}. Then, we get the predictions of in-between data through the models trained before, \textit{i.e.}, those presented   in Table \ref{tb2}, and calculate a prediction loss by measuring the cross-entropy loss between those predictions and soft labels. From the test results shown in Figure \ref{fig:test-inbet}, we could see that i) the cross-entropy loss on mixed data would become higher when the mix ratio $\lambda$ approaches 0.5, \textit{i.e.}, when making mixed bags far from both of two input bags; ii) it is clear that the models trained with PseMix often show lower test losses on out-of-distribution data points (\textit{i.e.}, $0.0<\lambda<1.0$). The second observation is consistent with our intuitive explanation to PseMix, \textit{i.e.}, PseMix-based models often perform better on mixed in-between data, suggesting that they have learned from new vicinity samples. 

\subsection{Special robustness tests}
\label{sec:rob}

Here we further demonstrate yet another benefit of our PseMix data augmentation scheme. TCGA-BRCA is used in this test. 

\subsubsection{Robustness against patch occlusion}
We randomly mask the instances of bags of test set samples and then use them for testing. From Table \ref{tb3}, we observe that PseMix-based models almost always obtain clear improvements over vanilla ones in various masking ratios. These improvements range from 1.92\% to 3.85\% and 2.57\% to 3.43\% in average ACC and AUC, apparently larger than those gains shown in Table \ref{tb2}. These facts suggest that our PseMix also could help models in the robustness to patch occlusion. 

\begin{table*}[htbp]
\centering
\caption{Different methods of pseudo-bag division. \textit{FT} indicates fine-tuning. $\overline{T}$ is the average time cost per slide for division.}
\label{tb7}
\begin{tabular}{l|l|cc|cc|cc|cc|c}
\toprule
\multirow{2}{*}{Network} & \multirow{2}{*}{Division method}  & \multicolumn{2}{c|}{TCGA-BRCA} & \multicolumn{2}{c|}{TCGA-LUNG} & \multicolumn{2}{c|}{TCGA-RCC} & \multicolumn{2}{c|}{Average (\%)} & $\overline{T}$ \\ 
 & & ACC (\%)        & AUC (\%)       & ACC (\%)        & AUC (\%)        & ACC (\%)       & AUC (\%) & ACC & AUC & ($\times {10}^{-2} s$)      \\ \midrule
\multirow{4}{*}{ABMIL} & random  & 88.01{\tiny\ $\pm$ 2.53} & 89.30{\tiny\ $\pm$ 3.26} & 86.06{\tiny\ $\pm$ 2.71} & 91.77{\tiny\ $\pm$ 2.16} & 89.55{\tiny\ $\pm$ 1.82} & 97.68{\tiny\ $\pm$ 0.46} & 87.87 & 92.92 & 0.016        \\
  & K-means & 86.97{\tiny\ $\pm$ 1.82} & 88.47{\tiny\ $\pm$ 2.35} & 85.21{\tiny\ $\pm$ 2.49} & 91.87{\tiny\ $\pm$ 1.92} & 89.95{\tiny\ $\pm$ 1.77} & 97.77{\tiny\ $\pm$ 0.46} & 87.38 & 92.70 & 154.329 \\
  & prototype & 86.33{\tiny\ $\pm$ 3.30} & 89.18{\tiny\ $\pm$ 3.36} & 86.06{\tiny\ $\pm$ 2.35} & 91.86{\tiny\ $\pm$ 2.20} & 89.66{\tiny\ $\pm$ 2.32} & 97.74{\tiny\ $\pm$ 0.46} & 87.35 & \textbf{92.93} & 0.129 \\
  & prototype + \textit{FT} & 87.50{\tiny\ $\pm$ 2.37} & 89.00{\tiny\ $\pm$ 3.74} & 86.32{\tiny\ $\pm$ 2.76} & 91.81{\tiny\ $\pm$ 2.40} & 90.08{\tiny\ $\pm$ 2.11} & 97.73{\tiny\ $\pm$ 0.47} & \textbf{87.97} & 92.85 & 0.382  \\ \midrule
\multirow{4}{*}{DSMIL}  &  random &  86.45{\tiny\ $\pm$ 1.76} & 89.33{\tiny\ $\pm$ 3.50} & 85.80{\tiny\ $\pm$ 3.19} & 93.12{\tiny\ $\pm$ 1.93} & 88.93{\tiny\ $\pm$ 1.26} & 97.60{\tiny\ $\pm$ 0.37} & 87.06 & 93.35 & 0.016 \\
   & K-means  & 84.45{\tiny\ $\pm$ 2.97} & 87.53{\tiny\ $\pm$ 4.11} & 86.85{\tiny\ $\pm$ 2.31} & 92.91{\tiny\ $\pm$ 1.26} & 88.92{\tiny\ $\pm$ 1.26} & 97.50{\tiny\ $\pm$ 0.54} & 86.74 & 92.65 & 154.329     \\ 
   & prototype  & 86.25{\tiny\ $\pm$ 2.33} & 89.57{\tiny\ $\pm$ 3.50} & 86.06{\tiny\ $\pm$ 3.79} & 92.65{\tiny\ $\pm$ 1.95} & 88.62{\tiny\ $\pm$ 2.18} & 97.41{\tiny\ $\pm$ 0.61} & 86.98 & 93.21 & 0.129     \\ 
   & prototype + \textit{FT} & 87.28{\tiny\ $\pm$ 1.83} & 89.72{\tiny\ $\pm$ 3.65} & 85.91{\tiny\ $\pm$ 2.77} & 92.89{\tiny\ $\pm$ 1.60} & 89.56{\tiny\ $\pm$ 2.38} & 97.64{\tiny\ $\pm$ 0.54} & \textbf{87.58} & \textbf{93.42} & 0.382    \\ \midrule
\multirow{4}{*}{TransMIL} & random &  86.67{\tiny\ $\pm$ 1.93} & 88.40{\tiny\ $\pm$ 2.37} & 84.31{\tiny\ $\pm$ 2.53} & 91.90{\tiny\ $\pm$ 1.74} & 90.53{\tiny\ $\pm$ 1.97} & 97.82{\tiny\ $\pm$ 0.19} & 87.17 & 92.71 & 0.016   \\
  & K-means & 87.31{\tiny\ $\pm$ 1.80} & 89.18{\tiny\ $\pm$ 2.48} & 85.71{\tiny\ $\pm$ 2.41} & 92.65{\tiny\ $\pm$ 2.24} & 89.89{\tiny\ $\pm$ 1.89} & 97.78{\tiny\ $\pm$ 0.33}  & 87.64 & \textbf{93.20} & 154.329     \\
  & prototype & 86.24{\tiny\ $\pm$ 2.32} & 88.78{\tiny\ $\pm$ 2.02} & 85.72{\tiny\ $\pm$ 1.54} & 92.09{\tiny\ $\pm$ 0.80} & 89.86{\tiny\ $\pm$ 2.81} & 97.75{\tiny\ $\pm$ 0.27} & 87.27 & 92.87 & 0.129     \\
  & prototype + \textit{FT} & 86.65{\tiny\ $\pm$ 2.26} & 88.50{\tiny\ $\pm$ 3.52} & 86.22{\tiny\ $\pm$ 1.47} & 92.63{\tiny\ $\pm$ 1.63} & 90.83{\tiny\ $\pm$ 1.08} & 97.78{\tiny\ $\pm$ 0.23}  & \textbf{87.90} & 92.97 & 0.382      \\ \bottomrule
%\multicolumn{8}{p{320pt}}{The overall best ones, i.e., the best sum of ACC and AUC, are in \textbf{bold}.}
\end{tabular}
\end{table*}

\begin{table*}[htbp]
\centering
\caption{Different target mixing strategies. MR means mixing ratio.}
\label{tba1}
\begin{tabular}{l|l|cc|cc|cc|cc}
\toprule
\multirow{2}{*}{Network}    & \multirow{2}{*}{$\lambda$ of target mixing} & \multicolumn{2}{c|}{TCGA-BRCA} & \multicolumn{2}{c|}{TCGA-LUNG} & \multicolumn{2}{c|}{TCGA-RCC} & \multicolumn{2}{c}{Average (\%)} \\ %\cmidrule(lr){3-4} \cmidrule(lr){5-6} \cmidrule(lr){7-8}
 &                         & ACC (\%)        & AUC (\%)       & ACC (\%)        & AUC (\%)        & ACC (\%)       & AUC (\%) & ACC & AUC  \\ \midrule
\multirow{2}{*}{ABMIL} & instance MR  & 86.76{\tiny\ $\pm$ 1.85} & 87.99{\tiny\ $\pm$ 2.31} & 85.00{\tiny\ $\pm$ 2.75} & 91.67{\tiny\ $\pm$ 2.14} & 90.18{\tiny\ $\pm$ 1.32} & 97.68{\tiny\ $\pm$ 0.42} & 87.31 & 92.45  \\
  & pseudo-bag MR & 87.50{\tiny\ $\pm$ 2.37} & 89.00{\tiny\ $\pm$ 3.74} & 86.32{\tiny\ $\pm$ 2.76} & 91.81{\tiny\ $\pm$ 2.40} & 90.08{\tiny\ $\pm$ 2.11} & 97.73{\tiny\ $\pm$ 0.47} & \textbf{87.97} & \textbf{92.85}  \\ \midrule
\multirow{2}{*}{DSMIL}  &  instance MR &  85.83{\tiny\ $\pm$ 2.07} & 88.49{\tiny\ $\pm$ 3.39} & 86.57{\tiny\ $\pm$ 2.97} & 92.83{\tiny\ $\pm$ 2.10} & 89.99{\tiny\ $\pm$ 2.70} & 97.46{\tiny\ $\pm$ 0.67} & 87.46 & 92.93 \\
   & pseudo-bag MR  & 87.28{\tiny\ $\pm$ 1.83} & 89.72{\tiny\ $\pm$ 3.65} & 85.91{\tiny\ $\pm$ 2.77} & 92.89{\tiny\ $\pm$ 1.60} & 89.56{\tiny\ $\pm$ 2.38} & 97.64{\tiny\ $\pm$ 0.54} & \textbf{87.58} & \textbf{93.42}  \\  \midrule
\multirow{2}{*}{TransMIL} & instance MR &  87.28{\tiny\ $\pm$ 1.88} &  89.05{\tiny\ $\pm$ 2.97} &  85.62{\tiny\ $\pm$ 2.06} &  92.94{\tiny\ $\pm$ 2.26} &  90.17{\tiny\ $\pm$ 2.09} &  97.68{\tiny\ $\pm$ 0.33} & 87.69 & \textbf{93.23}  \\
  & pseudo-bag MR & 86.65{\tiny\ $\pm$ 2.26} & 88.50{\tiny\ $\pm$ 3.52} & 86.22{\tiny\ $\pm$ 1.47} & 92.63{\tiny\ $\pm$ 1.63} & 90.83{\tiny\ $\pm$ 1.08} & 97.78{\tiny\ $\pm$ 0.23}  & \textbf{87.90} & 92.97  \\ \bottomrule
\end{tabular}
\end{table*}

\subsubsection{Robustness against label corruption}

To evaluate the model's robustness to label noise, we use label-corrupted training samples to train MIL models. Specifically, some training samples are randomly selected and then tagged with any one random class label, following \cite{zhang2018mixup}. Results are reported on original test sets using the best (at the epoch with minimum validation loss) and the last models (at the last epoch). We specially report performance for both the best model and the last one to assess the performance gap between them, so as to study the negative effect of label-corrupted training data on model performance. By comparing that gap between a vanilla model and its PseMix-based counterpart, we could infer that the model with a smaller gap may be more resistant to label-corrupted samples and has a more stable training process, \textit{i.e.}, it is more robust to label-corrupted training samples.

As shown in Table \ref{tb4}, (1) we find that the models trained with our PseMix often show great performance gains, 7 out of 9 cases for the best epoch and 8 out of 9 cases for the last one. Moreover, in average performance, PseMix-based models consistently exceed vanilla ones by large margins (1.80\% $\sim$ 5.58\% in AUC and 1.09\% $\sim$ 9.96\% in ACC). Furthermore, (2) we notice that vanilla models often suffer from greater performance drops than PseMix-based ones, when changing the epoch from best to last. For instance, a vanilla ABMIL model drops by 6.94\% in accuracy when turning to use the last model, for a 20\% corruption ratio; whereas its PseMix-based counterpart only drops by 3.34\% in the same case. These observations tell us that our PseMix could often make model training more stable and help MIL models to be more robust to label noise in training.

\subsection{Ablation study}
\label{sec:abl}

%In this section, we first try different pseudo-bag division methods and examine their effects on model performance. Then we investigate the two critical designs of our PseMix. 

\subsubsection{Study on different pseudo-bag division methods}

As described in Section \ref{sec:pse}, different methods could be adopted to generate pseudo-bags, such as random and phenotype-stratified sampling. Here we compare our adopted method, \textit{i.e.}, bag-prototype-based clustering + phenotype fine-tuning (prototype + \textit{FT}), with the other three, random sampling, K-means-based and bag-prototype-based stratified sampling. We set $p=1$ for this study. Test results are presented in Table \ref{tb7}. 

These test results indicate that our prototype + \textit{FT} is competitive in four pseudo-bag division methods. Our evidences have three-fold. (1) prototype + \textit{FT} could often obtain better overall performances than the other three, especially in average ACC. (2) It takes three orders of magnitude less time than the K-means-based one that also utilizes stratified sampling. (3) Compared to random and bag-prototype-based division, it merely introduces moderate time costs for phenotype clustering or fine-tuning but performs better in most cases. 

From the results of Table \ref{tb7}, we notice that compared with our prototype + \textit{FT}, a K-means-based method often shows a negative influence on PseMix data augmentation. One possible reason behind this is that the initial clusters obtained by bag-prototype-based clustering could help to yield a clearer phenotype hierarchy than those obtained by K-means. A clearer phenotype hierarchy could lead to the pseudo-bags that are more consistent with their parent bag in phenotype distribution, contributing to the semantic alignment in target mixing and thus improving the end performance of PseMix, as discussed in Section \ref{sec:psediv}. 
Specifically, our initial clusters are determined based on the similarity measurement between instances and bag prototype, and this similarity measurement is often helpful for differentiating instance phenotype \cite{yang2023protodiv}. By contrast, classical K-means ignores the characteristic of pathology WSIs to select initial cluster centers by random sampling. Moreover, its performance is known to be sensitive to initial clusters. As a result, it often takes a significantly longer time for convergence and yields a phenotype hierarchy worse than prototype + \textit{FT}.

\begin{table*}[htbp]
\centering
\caption{Ablation study on our PseMix. PB indicates only using pseudo-bags for data augmentation. \textit{FT}, Mixup, and R-mix means our phenotype fine-tuning, pseudo-bag-based Mixup, and random mixing, respectively.}
\label{tb8}
\begin{tabular}{l|c|ccc|cc|cc|cc|cc}
\toprule
\multirow{2}{*}{Method}  & \multirow{2}{*}{PB}  & \multicolumn{3}{c|}{Component}  & \multicolumn{2}{c|}{TCGA-BRCA} & \multicolumn{2}{c|}{TCGA-LUNG} & \multicolumn{2}{c|}{TCGA-RCC} & \multicolumn{2}{c}{Average (\%)} \\ %\cmidrule(lr){2-3} \cmidrule(lr){4-5} \cmidrule(lr){6-7} \cmidrule(lr){8-9}
  & & \textit{FT} & Mixup  & R-mix & ACC (\%)        & AUC (\%)       & ACC (\%)        & AUC (\%)        & ACC (\%)       & AUC (\%) & ACC       & AUC     \\ \midrule
 \multicolumn{13}{l}{- ABMIL \textit{network}} \\ \cmidrule{1-13}
\multirow{1}{*}{Baseline} & \checkmark & & &  & 86.43{\tiny\ $\pm$ 2.70} & 87.23{\tiny\ $\pm$ 2.04} & 85.20{\tiny\ $\pm$ 2.82} & 92.84{\tiny\ $\pm$ 1.94} & 91.36{\tiny\ $\pm$ 0.81} & 97.82{\tiny\ $\pm$ 0.62} & 87.66 & 92.63   \\ \cmidrule{1-13}
\multirow{4}{*}{PseMix} & \checkmark & \checkmark &  &  & 85.71{\tiny\ $\pm$ 2.25} & 87.39{\tiny\ $\pm$ 2.38} & 85.11{\tiny\ $\pm$ 2.60} & 92.19{\tiny\ $\pm$ 1.87} & 90.38{\tiny\ $\pm$ 0.84} & 97.94{\tiny\ $\pm$ 0.74} & 87.07 & 92.51   \\
  & \checkmark &  & \checkmark & & 86.33{\tiny\ $\pm$ 3.30} & 89.18{\tiny\ $\pm$ 3.36} & 86.06{\tiny\ $\pm$ 2.35} & 91.86{\tiny\ $\pm$ 2.20} & 89.66{\tiny\ $\pm$ 2.32} & 97.74{\tiny\ $\pm$ 0.46} & 87.35 & 92.93  \\
  & \checkmark & \checkmark & \checkmark &     & 87.50{\tiny\ $\pm$ 2.37} & 89.00{\tiny\ $\pm$ 3.74} & 86.32{\tiny\ $\pm$ 2.76} & 91.81{\tiny\ $\pm$ 2.40} & 90.08{\tiny\ $\pm$ 2.11} & 97.73{\tiny\ $\pm$ 0.47} & \textbf{87.97} & 92.85    \\
  & \checkmark & \checkmark & \checkmark & \checkmark & 86.64{\tiny\ $\pm$ 3.11} & 89.49{\tiny\ $\pm$ 3.69} & 86.45{\tiny\ $\pm$ 3.28} & 93.01{\tiny\ $\pm$ 2.06} & 90.50{\tiny\ $\pm$ 1.66} & 98.02{\tiny\ $\pm$ 0.55} & 87.86 & \textbf{93.51}    \\ \midrule \midrule
  \multicolumn{13}{l}{- DSMIL \textit{network}} \\ \cmidrule{1-13}
\multirow{1}{*}{Baseline}  & \checkmark &  &  &  & 86.13{\tiny\ $\pm$ 2.56} & 87.97{\tiny\ $\pm$ 3.12} & 87.16{\tiny\ $\pm$ 3.00} & 94.07{\tiny\ $\pm$ 2.74} & 89.95{\tiny\ $\pm$ 2.08} & 97.76{\tiny\ $\pm$ 0.77} & 87.75 & 93.27  \\ \cmidrule{1-13}
\multirow{4}{*}{PseMix} & \checkmark & \checkmark & & & 85.72{\tiny\ $\pm$ 1.63} & 89.31{\tiny\ $\pm$ 2.83} & 86.96{\tiny\ $\pm$ 2.91} & 93.57{\tiny\ $\pm$ 2.49} & 90.50{\tiny\ $\pm$ 1.79} & 97.93{\tiny\ $\pm$ 0.48} & 87.73 & 93.60  \\
   & \checkmark &  & \checkmark & & 86.25{\tiny\ $\pm$ 2.33} & 89.57{\tiny\ $\pm$ 3.50} & 86.06{\tiny\ $\pm$ 3.79} & 92.65{\tiny\ $\pm$ 1.95} & 88.62{\tiny\ $\pm$ 2.18} & 97.41{\tiny\ $\pm$ 0.61} & 86.98 & 93.21  \\
   & \checkmark & \checkmark & \checkmark & & 87.28{\tiny\ $\pm$ 1.83} & 89.72{\tiny\ $\pm$ 3.64} & 85.91{\tiny\ $\pm$ 2.77} & 92.89{\tiny\ $\pm$ 1.60} & 89.56{\tiny\ $\pm$ 2.38} & 97.64{\tiny\ $\pm$ 0.54} & 87.58 & 93.42 \\ 
   & \checkmark & \checkmark & \checkmark & \checkmark & 88.22{\tiny\ $\pm$ 2.65} & 89.65{\tiny\ $\pm$ 3.19} & 88.68{\tiny\ $\pm$ 2.19} & 93.92{\tiny\ $\pm$ 2.19} & 90.62{\tiny\ $\pm$ 1.20} & 97.89{\tiny\ $\pm$ 0.44} & \textbf{89.17} & \textbf{93.82} \\ \midrule \midrule
   \multicolumn{13}{l}{- TransMIL \textit{network}} \\ \cmidrule{1-13}
\multirow{1}{*}{Baseline} & \checkmark & & & & 86.66{\tiny\ $\pm$ 1.10} & 89.51{\tiny\ $\pm$ 1.29} & 87.50{\tiny\ $\pm$ 2.88} & 93.90{\tiny\ $\pm$ 1.79} & 89.34{\tiny\ $\pm$ 1.71} & 97.65{\tiny\ $\pm$ 1.05} & 87.83 & 93.69 \\ \cmidrule{1-13}
\multirow{4}{*}{PseMix} & \checkmark & \checkmark & & & 86.56{\tiny\ $\pm$ 0.91} & 89.31{\tiny\ $\pm$ 1.22} & 87.13{\tiny\ $\pm$ 3.37} & 93.33{\tiny\ $\pm$ 1.61} & 90.16{\tiny\ $\pm$ 1.82} & 97.69{\tiny\ $\pm$ 0.94} & 87.95 & 93.44 \\
  & \checkmark &  & \checkmark & & 86.24{\tiny\ $\pm$ 2.32} & 88.78{\tiny\ $\pm$ 2.02} & 85.72{\tiny\ $\pm$ 1.54} & 92.09{\tiny\ $\pm$ 0.80} & 89.86{\tiny\ $\pm$ 2.81} & 97.75{\tiny\ $\pm$ 0.27} & 87.27 & 92.87 \\
  & \checkmark & \checkmark & \checkmark &  & 86.65{\tiny\ $\pm$ 2.26} & 88.50{\tiny\ $\pm$ 3.52} & 86.22{\tiny\ $\pm$ 1.47} & 92.63{\tiny\ $\pm$ 1.63} & 90.83{\tiny\ $\pm$ 1.08} & 97.78{\tiny\ $\pm$ 0.23} & 87.90 & 92.97  \\
  & \checkmark & \checkmark &  \checkmark & \checkmark & 86.98{\tiny\ $\pm$ 1.47} & 90.40{\tiny\ $\pm$ 2.29} & 87.67{\tiny\ $\pm$ 2.80} & 93.47{\tiny\ $\pm$ 1.91} & 91.14{\tiny\ $\pm$ 1.94} & 97.76{\tiny\ $\pm$ 0.69} & \textbf{88.60} & \textbf{93.88} \\ \bottomrule
%\multicolumn{9}{p{390pt}}{The overall best ones, i.e., the best sum of ACC and AUC, are in \textbf{bold}. Tuning means fine-tuning phenotype clusters. Random means randomly taking pseudo-bags, rather than mixed bags, as training data.}
\end{tabular}
\end{table*}

\begin{table*}[htbp]
\centering
\caption{Different patch feature extractors.}
\label{tb5}
\begin{tabular}{l|l|cc|cc|cc|cc}
\toprule
\multirow{2}{*}{\makecell[l]{SSL-based\\ feature extractor}}    & \multirow{2}{*}{Method} & \multicolumn{2}{c|}{ABMIL} & \multicolumn{2}{c|}{DSMIL} & \multicolumn{2}{c|}{TransMIL} & \multicolumn{2}{c}{Average (\%)} \\ %\cmidrule(lr){3-4} \cmidrule(lr){5-6} \cmidrule(lr){7-8}
                          &                         & ACC (\%)        & AUC (\%)       & ACC (\%)        & AUC (\%)        & ACC (\%)       & AUC (\%) & ACC & AUC  \\ \midrule
\multirow{2}{*}{HIPT}    & vanilla                 & 84.89{\tiny\ $\pm$ 2.44} & 86.21{\tiny\ $\pm$ 2.22} & 82.50{\tiny\ $\pm$ 2.04} & 83.31{\tiny\ $\pm$ 2.88} & 82.80{\tiny\ $\pm$ 2.48} & 83.88{\tiny\ $\pm$ 3.55} & 83.40 & 84.47 \\
                          & w/ PseMix               & \textbf{85.09}{\tiny\ $\pm$ 2.39} & \textbf{86.39}{\tiny\ $\pm$ 2.85} & \textbf{85.50}{\tiny\ $\pm$ 2.82} & \textbf{87.00}{\tiny\ $\pm$ 2.92} & \textbf{84.89}{\tiny\ $\pm$ 2.06} & \textbf{84.94}{\tiny\ $\pm$ 3.44} & \textbf{85.16} & \textbf{86.11}          \\ \midrule
%                          & $\Delta$               & \textcolor[RGB]{36,141,88}{+ 0.20} & \textcolor[RGB]{36,141,88}{+ 0.18} & \textcolor[RGB]{36,141,88}{- 1.08} & \textcolor[RGB]{36,141,88}{+ 1.64} & \textcolor[RGB]{36,141,88}{+ 1.56} & \textcolor[RGB]{36,141,88}{+ 1.64}            \\ \midrule
\multirow{2}{*}{CTransPath}    & vanilla                 & 90.02{\tiny\ $\pm$ 1.99} & 92.83{\tiny\ $\pm$ 1.75} & 88.77{\tiny\ $\pm$ 1.74} & 92.58{\tiny\ $\pm$ 1.59} & 90.44{\tiny\ $\pm$ 1.46} & 92.97{\tiny\ $\pm$ 1.42} & 89.74 & 92.79 \\
                          & w/ PseMix               & \textbf{90.85}{\tiny\ $\pm$ 1.56} & \textbf{93.66}{\tiny\ $\pm$ 2.11} & \textbf{89.28}{\tiny\ $\pm$ 1.44} & \textbf{92.62}{\tiny\ $\pm$ 2.18} & \textbf{91.38}{\tiny\ $\pm$ 1.59} & \textbf{93.54}{\tiny\ $\pm$ 2.02} & \textbf{90.50} & \textbf{93.27}       \\
%                          & $\Delta$               & \textcolor[RGB]{36,141,88}{+ 0.83} & \textcolor[RGB]{36,141,88}{+ 0.83} & \textcolor[RGB]{36,141,88}{+ 0.62} & \textcolor[RGB]{36,141,88}{- 0.07} & \textcolor[RGB]{36,141,88}{+ 0.52} & \textcolor[RGB]{36,141,88}{+ 0.28}           \\
 \bottomrule
\end{tabular}
\end{table*}

\subsubsection{Study on different target mixing strategies}

Apart from the semantic alignment strategy based on pseudo-bag mixing ratio (called pseudo-bag MR), another strategy based on instance mixing ratio (called instance MR) could also be used for target mixing. Specifically, in $\widetilde{y}=\lambda y_A + (1-\lambda) y_B$, $\lambda$ could be determined according to the ratio of the number of instances from $A$ and $B$ in a mixed bag $\widetilde{X}$, instead of the ratio of the number of pseudo-bags. We set $p=1$ in this experiment to avoid the potential influence of random mixing.

From the comparative results of Table \ref{tba1}, we could see that the strategy of pseudo-bag MR leads to better overall performance than its instance MR-based counterpart, in 5 of 6 comparisons. These empirical results suggest that using a pseudo-bag MR as the parameter of target mixing ($\lambda$) could be better for semantic alignment than using an instance MR. This is largely caused by the characteristic of pseudo-bags, namely, pseudo-bags could inherit labels from their parent bag, which is the basic assumption of pseudo-bag. Due to this basic assumption, pseudo-bag is more likely to be aligned with its parent bag in semantics, compared with instance. Accordingly, utilizing the pseudo-bag MR for target mixing could often result in a better alignment of $\widetilde{X}$ with $\widetilde{y}$ in semantics. 

\subsubsection{Study on our PseMix}

On the basis of prior pseudo-bag-based methods, PseMix could be decomposed into these new components: a phenotype fine-tuning technique for pseudo-bag division (termed \textit{FT}), pseudo-bag Mixup (termed Mixup), and a random mixing mechanism (termed R-mix). To elucidate the influence of each new component on end performance, we conduct an ablation study on PseMix. The baseline of this ablation study is a pseudo-bag augmentation method (termed PB). It generates pseudo-bags by prototype-based clustering and stratified sampling, without \textit{FT} and Mixup.

From the ablation study results in Table \ref{tb8}, we have three empirical findings as follows. (1) For \textit{FT}, it has no obvious positive effects on pseudo-bag augmentation in terms of overall performance. However, its effectiveness on pseudo-bag Mixup is observed in 5 of 6 comparisons. These results show that pseudo-bag Mixup augmentation is more likely to benefit from phenotype fine-tuning than pseudo-bag augmentation. (2) For pseudo-bag Mixup augmentation, it is comparable with or slightly better than pseudo-bag augmentation (when using ABMIL) in presence of \textit{FT}, seen from the comparison between PB + \textit{FT} + Mixup and PB + \textit{FT}. A significant performance improvement is not often observed. This result indicates that a simple and direct mixing of pseudo-bags could not often lead to better performance, reflecting the challenge of applying Mixup to heterogeneous WSI bags.
(3) For our R-mix, we find that it obtains the best overall performance in 5 of 6 comparisons. This result suggests that utilizing both mixed and masked bags for data augmentation could often be more effective for MIL training than only using mixed bags (\textit{i.e.}, pseudo-bag Mixup augmentation) or masked bags (\textit{i.e.}, pseudo-bag augmentation). This result also demonstrates that our random mixing plays a critical role in PseMix. There are two possible reasons for this. The one is that random mixing brings more data diversity because it produces two kinds of augmented bags for training, \textit{i.e.}, mixing and masked bags, as described in Section \ref{sec-bag-mixing}. The other one is that the involvement of masked bags could make models learn from mixed bags (vicinity samples) more efficiently. Because pseudo-bag-masked bags could be cast as the intermediate samples between original training data and synthetic mixed data. These intermediate samples may enable models to learn from vicinity samples more efficiently, by smoothing the transition from original distribution to vicinity distribution. 

\subsection{Further analysis}
\label{sec:fur}

Here we examine more experimental settings to understand their sensitivity to model performance. We mainly present the experimental results measured on TCGA-BRCA. 

\subsubsection{Other feature extractors}

Apart from the classical ResNet-50 model adopted in \cite{Lu2021}, we also try two more advanced feature extractors. They are HIPT \cite{Chen2022} and CTransPath \cite{WANG2022102559}, pre-trained with self-supervised learning on WSIs. We directly use their released model weights for feature extraction. The test results of Table \ref{tb5} reveal that PseMix still consistently improves the performance of vanilla models across three MIL networks, even for different feature extractors. This result suggests the adaptability of PseMix. 

\subsubsection{Hyper-parameter analysis}
\begin{figure}[htbp]
\centering
\includegraphics[width=\columnwidth]{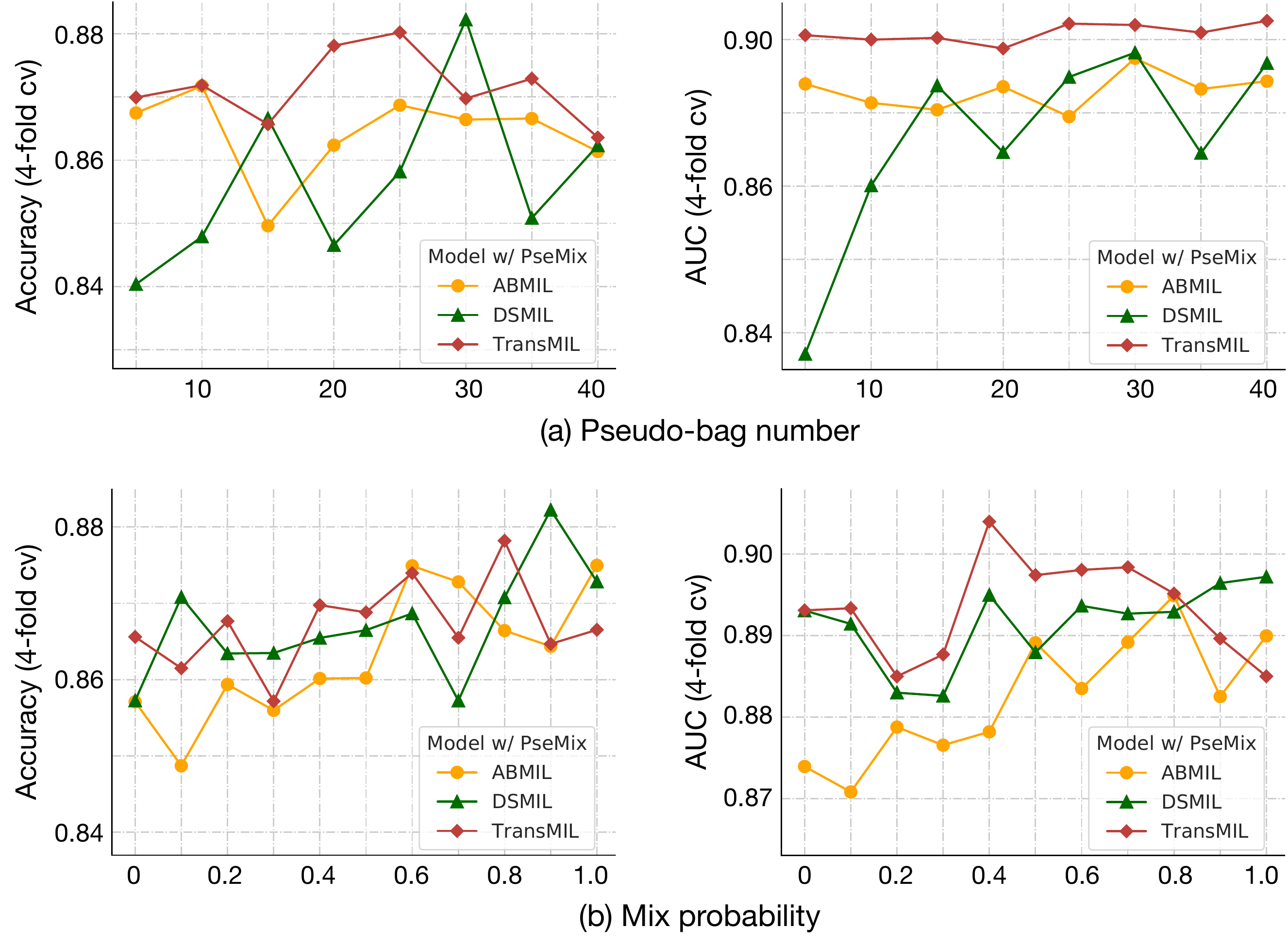}
\caption{Study on the two important hyper-parameters of PseMix, pseudo-bag number ($n$) and mix probability ($p$).}
\label{fig:test-params}
\end{figure}

We test the two most important hyper-parameters in PseMix, $n$ (the pseudo-bag number of each WSI bag) and $p$ (the probability of our random mixing). Test results are shown in Figure \ref{fig:test-params}. For $n$, a value near our default setting ($n=30$) tends to obtain better overall performance. All the hyper-parameters of pseudo-bag division are set by default across all datasets and networks without tuning. For $p$, a larger value is likely to result in better test performances. We set $p$ to 0.8, 0.9, and 0.4 for ABMIL, DSMIL, and TransMIL, respectively, based on their respective losses on validation sets; the other values of $p$ could obtain better results on test sets. 

\subsubsection{Combining with basic data augmentation}

Since some basic image augmentation methods could also be adopted for WSI data, we combine our PseMix with them to further examine its adaptability. Specifically, we follow \cite{Kang_2023_CVPR} to adopt two basic image augmentation methods, vertical flip and stain augmentation, for WSI patches. In MIL training, the patch features of each WSI are augmented by them with a probability of 50\%, in which each basic method is chosen with an equal probability. The model trained with basic augmentation methods is the baseline used for comparison.
From the results of Table \ref{tba2}, we find that our PseMix augmentation still improves the performance of MIL models in the presence of basic image augmentation. This experiment could further verify the adaptability of PseMix.

\begin{table}[t]
\centering
\caption{Combining PseMix with basic data augmentation methods.}
\label{tba2}
\begin{tabular}{l|l|cc}
\toprule
Network & Data augmentation & ACC (\%) & AUC (\%) \\ \midrule
\multirow{2}{*}{ABMIL} & baseline & 85.70{\tiny\ $\pm$ 2.20} & 88.35{\tiny\ $\pm$ 2.80} \\
 & baseline + PseMix & \textbf{87.49}{\tiny\ $\pm$ 2.32} & \textbf{89.34}{\tiny\ $\pm$ 3.27} \\ \midrule
%                          & $\Delta$               & \textcolor[RGB]{36,141,88}{+ 0.20} & \textcolor[RGB]{36,141,88}{+ 0.18} & \textcolor[RGB]{36,141,88}{- 1.08} & \textcolor[RGB]{36,141,88}{+ 1.64} & \textcolor[RGB]{36,141,88}{+ 1.56} & \textcolor[RGB]{36,141,88}{+ 1.64}            \\ \midrule
\multirow{2}{*}{DSMIL} & baseline & 87.92{\tiny\ $\pm$ 1.01} & 90.02{\tiny\ $\pm$ 2.09} \\
 & baseline + PseMix & \textbf{88.86}{\tiny\ $\pm$ 2.12} & \textbf{90.04}{\tiny\ $\pm$ 3.39}  \\ \midrule
%                          & $\Delta$               & \textcolor[RGB]{36,141,88}{+ 0.83} & \textcolor[RGB]{36,141,88}{+ 0.83} & \textcolor[RGB]{36,141,88}{+ 0.62} & \textcolor[RGB]{36,141,88}{- 0.07} & \textcolor[RGB]{36,141,88}{+ 0.52} & \textcolor[RGB]{36,141,88}{+ 0.28}           \\
\multirow{2}{*}{TransMIL} & baseline & 85.09{\tiny\ $\pm$ 1.94} & 89.15{\tiny\ $\pm$ 2.78} \\
 & baseline + PseMix & \textbf{86.97}{\tiny\ $\pm$ 1.77} & \textbf{89.56}{\tiny\ $\pm$ 3.32} \\ \bottomrule
\end{tabular}
\end{table}

\section{Conclusion}
This paper proposes a Pseudo-bag Mixup (PseMix) data augmentation scheme for MIL-based WSI classification. It utilizes a pseudo-bag concept to fulfill Mixup alignment, thereby generalizing the basic idea of Mixup from common images to special WSIs. This scheme is compatible with most prevalent MIL networks. Moreover, it is efficient and plugin-and-play, neither involving time-consuming operations nor relying on the prediction of MIL models. Comparative experiments and ablation studies confirm that PseMix is an effective Mixup variant for WSI classification. It could often improve the performance of MIL models and obtain better overall performance than other related mixing strategies. Other than that, it is observed that MIL models could often benefit more from PseMix in many other notable aspects, such as generalization gap, in-between data generalization, patch occlusion robustness, and label noise robustness. In the future, our PseMix could serve as a promising data augmentation method to help develop the WSI classification models with better generalization and robustness for clinical pathology diagnosis. 

\bibliographystyle{IEEEtran}
\bibliography{IEEEabrv,refs}

\end{document}